%% file: wacv_main_revision.tex
\documentclass[10pt,twocolumn,letterpaper]{article}

\usepackage{wacv}              

\usepackage[accsupp]{axessibility}
\usepackage{graphicx}
\usepackage{amsmath}
\usepackage{amssymb}
\usepackage{booktabs}
\usepackage{paralist}

\usepackage{mathtools}
\usepackage{bm}
\usepackage{pifont}
\usepackage{multirow,bigdelim}
\usepackage{subcaption}
\usepackage{listings}
\usepackage{amsfonts}
\usepackage{bbm}
\usepackage[ruled, vlined]{algorithm2e}
\usepackage{enumitem}
\usepackage{amsthm}

\usepackage[pagebackref,breaklinks,colorlinks]{hyperref}

\usepackage[capitalize]{cleveref}
\crefname{section}{Sec.}{Secs.}
\Crefname{section}{Section}{Sections}
\Crefname{table}{Table}{Tables}
\crefname{table}{Tab.}{Tabs.}

\newtheorem{assumption}{Assumption}
\newtheorem{lemma}{Lemma}
\newtheorem{definition}{Definition}

\newcommand{\DFRVAL}{DFR$_\text{Tr}^\text{Val}$~}
\newcommand{\ours}{TLDR}
\DeclareMathOperator*{\argmax}{arg\,max}

\def\wacvPaperID{681} 
\def\confName{WACV}
\def\confYear{2025}

\usepackage{imakeidx}

\begin{document}

\title{TLDR: Text Based Last-layer Retraining for Debiasing Image Classifiers}

\author{Juhyeon Park\textsuperscript{\rm 1}\thanks{Equal contribution},  Seokhyeon Jeong\textsuperscript{\rm 2}\footnotemark[1],  and Taesup Moon\textsuperscript{\rm 1, \rm 2, \rm 3}\thanks{Corresponding author} \\
{\tt\small \{parkjh9229, sh102201, tsmoon\}@snu.ac.kr}
\\
\textsuperscript{\rm 1} IPAI, Seoul National University 
\textsuperscript{\rm 2} ECE, Seoul National University \\
\textsuperscript{\rm 3} ASRI / INMC / AIIS, Seoul National University
} 
\maketitle
\input{contents/0.abstract}
\section{Introduction}
\input{contents/1.introduction}
\section{Preliminaries}
\input{contents/3.preliminaries}
\section{Method: \ours}
\input{contents/4.methods}

\section{Experimental Results}
\input{contents/5.experimental_results}
\section{Conclusion \& Limitations}
\input{contents/6.conclusion}
\section*{Acknowledgements}

This work was supported in part by the National Research Foundation of Korea (NRF) grant [No.2021R1A2C2007884] and by Institute of Information \& communications Technology Planning \& Evaluation (IITP) grants [RS-2021-II211343, RS-2021-II212068, RS-2022-II220113, RS-2022-II220959] funded by the Korean government (MSIT). It was also supported by AOARD Grant No. FA2386-23-1-4079. Juhyeon Park is grateful for financial support from Hyundai Motor Chung Mong-Koo Foundation.

{\small
\bibliographystyle{ieee_fullname}
\bibliography{main}
}

\newpage
\appendix
\onecolumn

\pagenumbering{gobble}
\setcounter{tocdepth}{2}
\tableofcontents

\clearpage
\pagenumbering{arabic} 
\setcounter{page}{1}   
\input{contents/supplementary}

\end{document}

%% file: contents/0.abstract.tex
\begin{abstract}
An image classifier may depend on incidental features stemming from a strong correlation between the feature and the classification target in the training dataset.
Recently, Last Layer Retraining (LLR) with group-balanced datasets is shown to be efficient in mitigating the spurious correlation of classifiers.
However, the acquisition of image-based group-balanced datasets is costly, which hinders the general applicability of the LLR method.
In this work, we propose to perform LLR based on text datasets built with large language models to debias a general image classifier.
To that end, we demonstrate that text can generally be a proxy for its corresponding image beyond the image-text joint embedding space, which is achieved with a linear projector that ensures orthogonality between its weight and the modality gap of the joint embedding space.
In addition, we propose a systematic validation procedure that checks whether the generated words are compatible with the embedding space of CLIP and the image classifier, which is shown to be effective for improving debiasing performance.
We dub these procedures as \ours \ (\textbf{T}ext-based \textbf{L}ast layer retraining for \textbf{D}ebiasing image classifie\textbf{R}s) and show our method achieves the performance that is competitive with the LLR methods that require group-balanced image dataset for retraining.
Furthermore, \ours \ outperforms other baselines that involve training the last layer without any group annotated dataset. Codes: \url{https://github.com/beotborry/TLDR}

\end{abstract}

%% file: contents/1.introduction.tex
\begin{figure}[t]
\centering
\includegraphics[width=\columnwidth, height=4cm]{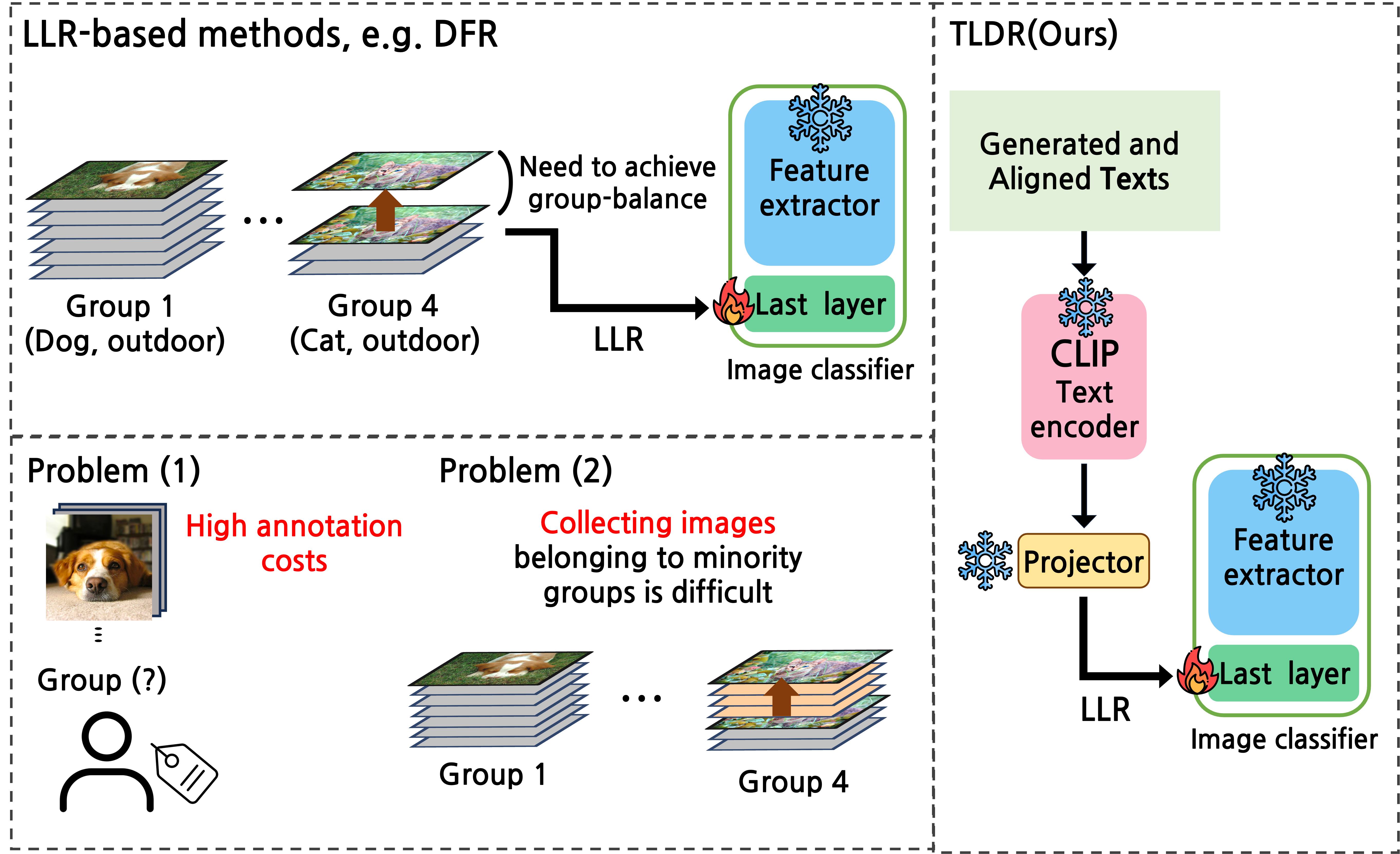}
\vspace{-0.2in}
\caption{
(Left) Typical procedure of LLR-based methods for debiasing an image classifier, which still suffers from high annotation cost for collecting group-balanced holdout image dataset, particularly when the number of classes and groups increases. 
(Right) In contrast, \ours \ effectively constructs a sufficient group-balanced set using only with \textit{texts}, by generating a substantial amount of data through LLMs with ease, 
to debias \textit{any} general image classifier.
More details are depicted in \cref{fig:main_method}.
}
\vspace{-0.2in}
\label{fig:intro}
\end{figure}

An image classifier may grant excessive importance to an inconsequential attribute of an input image as a result of detecting a strong correlation between the target and the attribute discovered in the training dataset. Such \textit{spurious} correlations can present a substantial problem in domains such as medical AI and autonomous driving, in which classification errors can cause severe consequences for humans.

To that end, numerous methods have been proposed to reduce the classifier's reliance on the spurious features, \textit{i.e.,} to \textit{debias} the classifier \cite{GDRO, JTT, SSA, CnC, SUBG, DFR}. Among those, DFR \cite{DFR} was recently proposed --- it suggests collecting a small holdout dataset that has a balanced number of data samples for each group, which stands for a sub-dataset with the same spurious attribute and target labels, and only retrain the last linear classification layer of a biased classifier based on the collected group-balanced holdout data. Such a procedure is often dubbed as Last Layer Re-training (LLR) and is shown to be very effective in debiasing a classifier. However, such an approach of collecting a group-balanced holdout dataset has a few critical limitations as shown in \cref{fig:intro}. Namely, the size of the holdout set still needs to be considerable to ensure the debiasing (namely, cannot be in the regime of \textit{few-shot} setting), hence, the annotation cost becomes expensive. Moreover, collecting a group-\textit{balanced} dataset becomes significantly more difficult as data pertaining to the minority group is often scarce and hard to collect in the wild. Furthermore, as the number of classes or spurious attributes increases, the cost of collecting a group-balanced image dataset increases exponentially.

In order to lift the burden of the requirement of collecting additional group-balanced datasets, DrML \cite{DRML} recently proposed a way to retrain the last linear layer of an image classifier only with \textit{text} data, thanks to the availability of the powerful joint embedding space produced by a multimodal pre-trained encoder, such as CLIP \cite{CLIP}.
Namely, for an image classifier that consists of a linear layer operating on top of the CLIP image embedding space, they showed that it is possible to achieve \textit{cross-modal transferability} --- \textit{i.e., } the text embedding can be used as a proxy for its corresponding image embedding in CLIP embedding space. Based on this finding, they used the text data to retrain the biased linear layer and showed the resulting debiased classifier still can work well with the image embeddings. 

However, we argue their result was limited since the proposed method is only applicable to the image classifiers that are defined on top of the joint embedding space. To the best of our knowledge, whether the cross-modal transferability can be extended to a more general embedding space, \textit{e.g.,} image embedding space obtained by a ResNet classifier, and hence, whether the text can be used to debias a general image classifier has not been investigated yet. 
In addition, DrML solely relied on the additional metadata for each benchmark dataset to generate the text dataset, which again limits the applicability to a general setting in which such metadata is not available.
Furthermore, using only the metadata to generate the text dataset may not be sufficient to achieve optimal debiasing performance due to its limited quantity or incompatibility with the embedding space of CLIP.


To that end, we aim to address the limitations of both DFR \cite{DFR} and DrML \cite{DRML} to develop a framework for debiasing a \textit{general} image classifier solely using text. Namely, unlike DFR, we remove the requirement of collecting annotated group-balanced \textit{image} data by developing general mechanisms for generating text 
data, without using any metadata. Moreover, unlike DrML, we achieve cross-modal transferability across different embedding spaces so that our method can be applied to debiasing general image classifiers.
Specifically, we first develop a linear projector that can project a CLIP embedding to another image embedding space while preserving the cross-modal transferability, considering the concept of the modality gap \cite{MindTheGap} of CLIP joint embedding space. Then, only with the knowledge of the existence and the type of spurious correlations, we \textit{generate} group-balanced text data using publicly available Large Language Models (LLM), such as GPT \cite{GPT} or LLaMA\cite{Llama}. 
The generated texts, which are not confined to some limited metadata, are then followed by a systematic procedure to only select those that are appropriate for debiasing. 
Finally, we retrain the last layer of the biased image classifier using the projected CLIP text embeddings of the generated group-balanced text data. In our experimental results, we show our method, dubbed as \ours \ (\textbf{T}ext-based \textbf{L}ast layer retraining for \textbf{D}ebiasing image classifie\textbf{R}s), outperforms other state-of-the-art LLR-based methods that do not require group-annotated data as ours and is competitive with DFR \cite{DFR}, which additionally uses the expensive group-annotated, group-balanced \textit{image} dataset.

Our contributions are summarized as follows.
\begin{itemize}[nosep,leftmargin=1em,labelwidth=*,align=left]
  \sloppy
  \item We propose a sufficient condition for preserving cross-modal transferability within the general image classifier's embedding space when linear alignment between embedding spaces is possible. 
  Moreover, we present a closed-form solution of a linear projector between embedding spaces with the condition, which eliminates the need for gradient-based optimization to learn the projector.
  \item We introduce a systematic process to construct a sufficient number of text data for LLR and select only those that are well-aligned with the embedding spaces of CLIP and the given image classifier.

  \item We experimentally demonstrate that our TLDR achieves competitive performance in debiasing a general image classifier and show it is particularly effective when the minority group has a considerably low data proportion. 
\end{itemize}

%% file: contents/3.preliminaries.tex
\subsection{Problem Setting}
Our work shares the group robustness problem setting introduced in \cite{GDRO}.
The data distribution can be specified by the group $\mathcal{G}$ defined as the Cartesian product of a set of labels $\mathcal{Y}$ and a set of spurious attributes $\mathcal{A}$, \textit{i.e., }$\mathcal{G} \coloneqq \mathcal{Y} \times \mathcal{A}$.
For example, in the Waterbirds dataset \cite{GDRO}, the label indicates whether a bird in an image is a landbird or waterbird, and the spurious attribute is the background (BG) of the image.
Thus, the group can be specified as $\mathcal{G} = $ \{landbirds, waterbirds\} $\times$ \{land BGs, water BGs\}.
Due to the prevalence of waterbirds with water BGs as well as landbirds with land BGs, the minority groups are (landbirds, water BGs) and (waterbirds, land BGs).
The reliance of the classifier on the spurious features is typically evaluated by the Worst Group Accuracy (WGA).

\subsection{Prior Works on Group Robustness}
Plenty of works have been suggested to mitigate spurious correlation problems in classification \cite{GDRO, JTT, SSA, CnC, SUBG, DFR}.
Existing works can be categorized with the level of assumption on the group information.
For instance, when the group information on the whole dataset is fully available, we can exploit such information to achieve group robustness, as in Group DRO \cite{GDRO}, by minimizing the worst group loss. 
As mentioned in the Introduction, our work is closely aligned with studies that use LLR (or fine-tuning) with holdout dataset, often dubbed as \textit{reweighting dataset}. We describe the most directly related work below, and provide a more comprehensive review in Sec. A of the Supplementary Materials (S.M.).


\noindent \textbf{DFR} \cite{DFR} reveals that deep neural networks often persist in learning core features despite the existence of spurious correlations in the training dataset. 
Consequently, they demonstrated that simple LLR with a group-balanced dataset alone can achieve group robustness.
This method is cost-effective and less complex as it does not require retraining of the entire classifier.
However, as pointed out earlier, collecting the group-annotated and group-balanced image dataset can often become costly, particularly when the number of groups increases and the samples from the minority group are hard to collect. 
 
\noindent
\textbf{AFR} \cite{AFR} merges the concept of the LLR and the inference of group information of the data \cite{JTT, SSA, CnC}. 
They proposed LLR of the Empirical Risk Minimization (ERM) model with a weighted loss function that assigns higher importance to instances where the ERM model exhibits poor predictions, thereby prioritizing the minority group. 
However, their approach necessitates a split of the original training dataset to create a reweighting dataset for LLR, potentially requiring additional training of the ERM model even when a pre-trained model is available.
Furthermore, as the proportion of the minority group decreases, the performance may significantly drop as the reweighting dataset may not contain enough minority examples.


\noindent
\textbf{SELF} \cite{SELF} leverages training checkpoints akin to \cite{JTT} to infer data belonging to a minority group in a holdout dataset, constructed from half of the validation set.
Specifically, they select data where the prediction discrepancies between the fully trained model and the early stopped model are substantial, then these selected data constitute of the reweighting dataset used for fine-tuning the last layer.
This process involves additional training of the entire model to store checkpoints of the early stopped model and to perform class-balanced ERM, which cannot utilize the ERM-trained model.
Furthermore, if the proportion of minority examples in the class-balanced ERM training phase is low, the performance of the model may decline because the disagreement-based inference of the group information can be inaccurate.
\subsection{Notations}
We denote the feature extractor and the last linear layer of a general image classifier by $f_\theta$ and $h_\phi$, with parameters $\theta$ and $\phi$ respectively. We mainly consider two embedding spaces --- a \textit{joint} embedding space generated by image-text contrastive learning-based models, \textit{e.g.,} CLIP \cite{CLIP} or ALIGN \cite{ALIGN}, and an image embedding space generated by the penultimate layer of a \textit{general} image classifier. The representation vector for each space and data is explicitly denoted by $\bm z_{\text{modality}}^{\text{space}}$; \textit{e.g.}, an embedding of text $T$ residing in the CLIP's embedding space is represented as $\bm z_{T}^{\text{CLIP}}\in\mathbb{R}^{d_{\text{CLIP}}}$ and an embedding of an image $I$ obtained by $f_\theta$ is denoted by  $\bm z_{I}^{f_\theta}\in\mathbb{R}^{d_{f_\theta}}$. Note $d_{\text{CLIP}}$ and $d_{f_{\theta}}$ denote the dimension of each embedding space. We mainly consider  CLIP \cite{CLIP} as a representative 
joint embedding space throughout the paper.



\subsection{Cross-modal Transferability}
\label{sec:cross_modal_transferability}

In \cite{MindTheGap}, the existence of the modality gap, which refers to a constant gap between image and text embeddings in the joint embedding space, was first discovered.
Following \cite{DRML}, we consider the instance-wise modality gap, of which the definition is restated below. 

\begin{definition}[Modality gap in CLIP]\label{def:gap}
Let $(I,T)$ denote an image-text pair. Then, the modality gap $\bm g$ in the joint embedding space of CLIP is defined  as \[\bm g \coloneqq \bm z^{\text{CLIP}}_{I} - \bm z^{\text{CLIP}}_{T}. \]
Empirically, $\bm g$ can be approximated as a constant regardless of $(I, T)$.
\end{definition}
\noindent \textit{Remark: }Although \cref{def:gap} is with respect to CLIP space, $\bm g$ can be defined in any joint embedding space where two modalities are well aligned \cite{MindTheGap}.


The existence of the modality gap does not affect the achievement of cross-modal transferability, which enables using embeddings from different modalities interchangeably. 
Namely, for an image-text pair $(I,T)$, \cite{DRML} showed that $h(\bm z_{I}^{\text{CLIP}}) \approx h(\bm z_{T}^{\text{CLIP}})$, in which $h$ is a linear classifier on top of the CLIP embedding space. However, the limitation of such cross-modal transferability is that it could be only achieved on the joint embedding space like CLIP and not on other general embedding spaces, \textit{e.g.,} an embedding space obtained by a general feature extractor $f_\theta$. 

In this paper, we aim to address such a limitation and enable the cross-modal transferability beyond of the joint embedding space. To that end, we consider a \textit{linear} projector $\bm \Pi:\mathbb{R}^{d_{\text{CLIP}}}\rightarrow \mathbb{R}^{d_{f_\theta}}$ that projects $\bm z_I^{\text{CLIP}}$ to $\bm z_{I}^{f_\theta}$. Namely, we denote $(\bm W, \bm b)$ as the linear matrix and bias vector that defines $\bm\Pi$. Then, we make the following (rough) assumption, which we believe is sensible because both $\bm z_{I}^{\text{CLIP}}$ and $\bm z_{I}^{f_\theta}$ reside in linearly separable spaces in the sense that the high average accuracy can be achieved with a linear layer.
\setcounter{assumption}{0}
\begin{assumption}\label{assumption}
    There exists a linear projector $\bm \Pi$ that makes $\bm \Pi(\bm z_{I}^{\text{CLIP}}) \approx \bm z_{I}^{f_\theta}$. More specifically, we assume $\|\bm \Pi(\bm z_I^{\text{CLIP}})-\bm z_I^{f_\theta}\|_2^2 < \|\bm z_{I_1}^{f_\theta}- \bm z_{I_2}^{f_\theta}\|_2^2$  for any arbitrary images $I_1$ and $I_2$ that belong to the same class as $I$. 
\end{assumption}
The discussion on whether the assumption is valid is presented in Sec. C.4 of the S.M.
Now, for the cross-modal transferability on the embedding space of $f_\theta$, we would like to achieve $h_\phi(\bm z_{I}^{f_\theta}) \approx h_\phi(\bm \Pi(\bm z_{T}^{\text{CLIP}}))$ for a pair $(I,T)$, namely, we would like to use the projected embedding of CLIP text embedding interchangeably for the image embedding on $f_\theta$. We can then derive a \textit{sufficient condition} for the projector $\bm \Pi$ under our assumption by examining the equations below:
\begin{align}
    &h_\phi(\bm\Pi(\bm z_{T}^{\text{CLIP}})) = 
    h_\phi(\bm W^\top \bm z_{T}^{\text{CLIP}} + \bm b)  \\
    &= h_\phi(\bm W^\top (\bm z_{I}^{\text{CLIP}} - \bm g) + \bm b) \label{eq:assump1_eq_1} \\
    &= h_\phi( \bm \Pi(\bm z_{I}^{\text{CLIP}}) - \bm W^\top \bm g) \approx h_\phi(\bm z_{I}^{f_\theta} - \bm W^\top \bm g), \label{eq:assump1_eq_2}
\end{align}
in which \cref{eq:assump1_eq_1} follows from \cref{def:gap}, and \cref{eq:assump1_eq_2} follows from the assumption and the continuity of $h_\phi$. Thus, from the above equations, we can easily deduce the sufficient condition for the cross-modal transferability is $\bm W^\top \bm g=0$; \textit{i.e.,} the modality gap $\bm g$ should lie in the nullspace of $\bm W^\top$. Guided by this sufficient condition, we obtain our linear projector $\bm \Pi$ and introduce our method in detail in the next section. 

%% file: contents/4.methods.tex
\begin{figure*}[t]
\includegraphics[width=\textwidth, height=6cm]{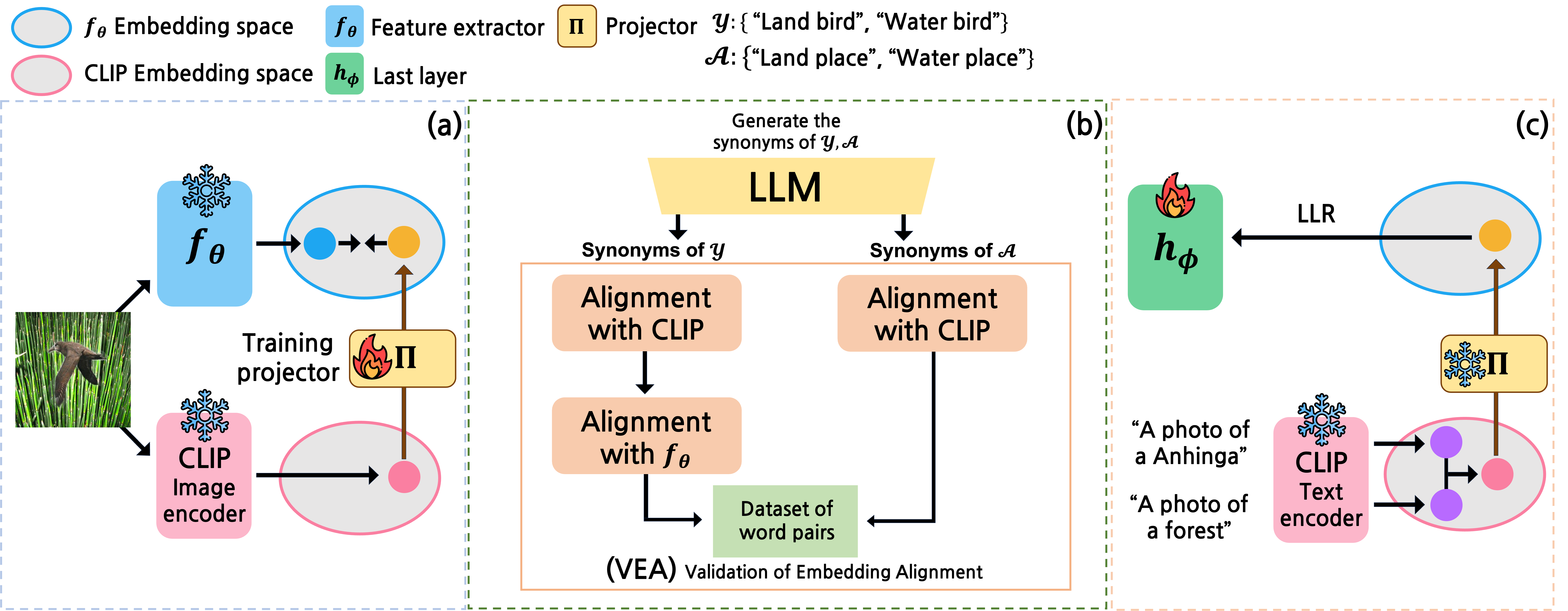} 
\vspace{-0.25in}
\caption{ 
\textit{(a)} Diagram of obtaining the projector $\bm\Pi$. The same image is fed to the encoder of $f_\theta$ and CLIP image encoder. Then, $(\bm W^*, \bm b^*)$ of $\bm \Pi$ minimizes the distance between the embedding of $f_\theta$ and the projected embedding from CLIP.
\textit{(b)} Diagram of the validation of the embedding alignment (VEA) of the generated words.
\textit{(c)} Diagram of the text-based LLR. Text embeddings for each class and spurious attribute are averaged in the embedding space of CLIP and projected to the embedding space of $f_\theta$ with $\bm\Pi$. Consequently, the projected embedding is fed to $h_\phi$, which is used for retraining.
}
\label{fig:main_method}
\vspace{-0.2in}
\end{figure*}


Our TLDR  is summarized as follows and in \cref{fig:main_method}.
\begin{enumerate}[nosep,leftmargin=1em,labelwidth=*,align=left]
    \item Estimating $(\bm W,\bm b)$ of $\bm \Pi$ which connects embedding spaces generated by the CLIP and a trained image classifier. (\cref{sec:proj}, \cref{fig:main_method}(a))
    \item Generating synonyms for category names of classes and spurious attributes with LLM and validating whether the generated words are well aligned with the embedding space of CLIP and the trained image classifier. (\cref{sec:filter}, \cref{fig:main_method}(b))
    \item Text-based LLR with the projected embedding of the text prompts, which are obtained from the words validated by Step 2. (\cref{sec:llr}, \cref{fig:main_method}(c))
\end{enumerate}

\subsection{Closed-Form Estimation of the Projector} 
\label{sec:proj}
In line with \cref{lemma_proj}, we impose the constraint $\bm W^\top \bm g = 0$ on $\bm \Pi$ to extend the cross-modal transferability in an embedding space of an arbitrary image classifier.
We simply estimate $\bm g$ by sampling image-text pairs from the COCO-Caption dataset \cite{CoCoCaption} and averaging their gaps, \textit{i.e.,} $\hat{\bm g} = \frac{1}{N} \sum_{i=1}^{N} (\bm z_{I_i}^{\text{CLIP}} - \bm z_{T_i}^{\text{CLIP}})$.
We emphasize that the gap estimates can be easily obtained from open-sourced image-text paired dataset \cite{CoCoCaption, LAION} and the estimated gap is independent of the dataset used to train the image classifier, $f_\theta$.
With the estimated gap, we solve the constrained ridge regression problem to obtain $(\bm W, \bm b)$ of $\bm \Pi$.
Even if there is a discrepancy between $\hat {\bm g}$ and each $(\bm z_{I_i}^{\text{CLIP}} - \bm z_{T_i}^{\text{CLIP}})$, the impact is minimal in our method, as detailed in Sec. C.5 of the S.M.


\begin{lemma}[Constrained ridge regression estimate of $\bm \Pi$] 
\label{lemma_proj}

\noindent Let $X \in \mathbb{R}^{n \times d_{\text{CLIP}}}$ be a matrix of CLIP embeddings of $n$ training images, $Y \in \mathbb{R}^{n \times d_{f_\theta}}$ be a matrix of embeddings generated by $f_\theta$ for the corresponding images. Moreover, let $\bm{W} \in \mathbb{R}^{d_{\text{CLIP}} \times d_{f_\theta}}, \bm b \in \mathbb{R}^{d_{f_\theta}}$ define the linear projector $\bm \Pi$ and $\bm g \in \mathbb{R}^{d_{\text{CLIP}}}$ be a modality gap. Following Assumption \ref{assumption}, we assume the following linear relationship between $X$ and $Y$:
\begin{equation}
    Y \sim X\bm W + \bm b+ \bm \epsilon, \ \ \ \bm \epsilon \sim \mathcal{N}(\bm 0, \bm\Sigma).
\end{equation}
Then, the ridge regression estimate of $(\bm W,\bm b)$ with the constraint $\bm W^\top \bm g = 0$ is
\begin{equation}
    \bm W^* = \Tilde{\bm W} - (X^\top X+\lambda I)^{-1}\bm{g}(\bm{g}^\top (X^\top X+\lambda I)^{-1}\bm{g})^{-1}\bm{g}^\top \Tilde{\bm W}
\end{equation}
\begin{equation}
    \bm b^* = \frac{1}{n}(Y - X\bm W^*)^\top \mathbf{1}
\end{equation}
in which $\Tilde{\bm W} = (X^\top X+\lambda I)^{-1}X^\top Y$, and $\mathbf{1}$ is a all-ones vector. (We also assume $\bm\Sigma$ is diagonal.)



\end{lemma}
The proof of the lemma is in Sec. D of the S.M.
$\lambda$ is a hyperparameter for the $\ell_2$ regularization and searched based on the NMSE \cite{CounTEX} criterion, which evaluates the normalized $\ell_2$ distance between original embedding and its prediction, \textit{i.e., } $\text{NMSE}(\bm z, \bm{\hat{z}}) \coloneqq \frac{||\bm z-\bm {\hat{z}}||^2_2}{||\bm z||^2_2}$.
Throughout the paper, $(\bm W^*, \bm b^*)$ are computed with the training set of each dataset, and $\lambda$ is searched with the corresponding validation set, following the same procedure as training the image classifier.
In addition, we do not perform $\ell_2$ normalization on the CLIP embeddings and $\bm g$ turns out to be almost constant without any normalization; refer to Sec. C.2 of the S.M.

It is important to note that our proposed algorithm has a clear advantage in that it eliminates the need for gradient descent-based optimization for learning the $\bm \Pi$ and additional tedious search of hyperparameters. 
(\textit{e.g., } learning rate, weight decay, etc.)
Moreover, it turns out that training $\bm \Pi$ to minimize $||\bm z_I^{f_\theta} - \bm \Pi \bm z_T^{\text{CLIP}}||^2_2$ is not as effective as our approach since it does not sufficiently consider the domain in which the ($f_\theta$, $h_\phi$) are trained.
The detailed analysis on this is discussed in \cref{sec:pi_analysis}.
Besides, there are alternative methods for mitigating the impact of $\bm g$ such as adding $\hat {\bm g}$ to $\bm z_T^{\text{CLIP}}$ to recover $\bm z_I^{\text{CLIP}}$, or methods suggested by \cite{Icantbelieve, decap}.
However, we found that these are less effective than ours and have some limitations; refer to Sec. C.6 of the S.M.

\subsection{Text-based Dataset Construction}
\label{sec:filter}

Given the category names of the class label and spurious attribute in $\mathcal{Y}$ and $\mathcal{A}$, respectively, we generate diverse words corresponding to those names by prompting GPT-3.5 \cite{GPT} to generate synonyms of them. Such diversity is essential in carrying out LLR as we show in our experiments.
We denote the set of generated words for the $y$-th element in $\mathcal{Y}$ as $\mathcal{T}^y$ and the $i$-th generated word is denoted as $t^y_i \in \mathcal{T}^y$. 
$t^a_i \in \mathcal{T}^a$ is defined in the same way for $\mathcal{A}$.
The full list of generated words is given in Sec. E.8 of the S.M. 

Despite the impressive generation capability of LLMs, it is well-known that they can exhibit hallucinations and provide inaccurate information \cite{hallucination, hallucination_survey}. 
Furthermore, the generated texts and their corresponding projected embeddings may be out-of-distribution (OOD) examples for CLIP and $f_\theta$ respectively, which can diminish the performance of LLR.
Hence, we implement validation of embedding alignment (VEA) for the generated words to only keep those that are compatible with embedding spaces of CLIP and $f_\theta$. 

\noindent \textbf{[VEA for CLIP]}
Since we expect the CLIP embeddings of the generated words in $\mathcal{T}^y$ for each $y\in\mathcal{Y}$  to be close to $\bm z_y^{\text{CLIP}}$, we employ the following simple and effective rule to remove words that are misaligned with CLIP --- we only select $t_i^y$ of which cosine similarity with its corresponding class $y$ in the CLIP text embedding space is the largest. Namely, $t_i^y$ is selected only when
\begin{equation}
    \arg\max_{y'\in\mathcal{Y}}\ \ \cos \Big(\bm z_{P_1(t_i^y)}^{\text{CLIP}}, \bm z_{P_1(y')}^{\text{CLIP}} \Big) = y
 \end{equation}
holds. It is done for the generated words for $\mathcal{T}^a$ analogously.

\noindent \textbf{[VEA for $f_\theta$]}
Once the VEA for CLIP is done, the remaining words will be semantically well-aligned with the CLIP embedding space. 
However, particularly for the remaining words in $\mathcal{T}^y$, they still may not be well-aligned with the $f_\theta$'s embedding space since their ``projected'' CLIP embeddings may be OOD examples within the embedding space of $f_\theta$.
Such misalignment may cause additional confusion when building a new classification boundary via LLR. To that end, we implement a logit-based VEA step for $f_\theta$ to further remove the misaligned words in $\mathcal{T}^y$ --- namely, we only select $t_i^y$ that satisfies
\begin{equation}
    \argmax_{y'\in\mathcal{Y}} \ \ h_\phi\Big(\bm \Pi(\bm z^{\text{CLIP}}_{P_1(t^{y}_i)})\Big)_{y'} = y,
\end{equation}
in which $\bm\Pi$ is the projector defined by $(\bm W^*,\bm b^*)$, $h_\phi(\cdot)_{y'}$ denotes the $y'$-th element of $h_\phi(\cdot)$, and $P_1$ is a prompt template with $P_1(t) = $ \texttt{"A photo of a $\{t\}$"}. 
We note that this VEA step is \textit{only} done for the words in $\mathcal{T}^y$.

\noindent \textbf{[Constructing Dataset of Word Pairs]}
We construct a dataset of word pairs based on the remaining words after the VEA.
We simply consider all possible combinations for each group, \ie, $\mathcal{T}^y \times \mathcal{T}^a$ for group $(y, a)$ where $\mathcal{T}^y, \mathcal{T}^a$ now include only validated words.

We emphasize that our proposed word generation and VEA procedure enables text-based LLR in a situation where appropriate words are not given in the metadata of a dataset, which was not sufficiently discussed in DrML \cite{DRML}.
In addition, we effectively improve the performance of LLR by using a sufficient amount of text data compared to DrML, which is discussed in detail in \cref{sec:drml_compare}.

\subsection{Text-based Last Layer Retraining} 
\label{sec:llr}
We use the word pairs to compute the CLIP text embeddings, which are projected onto the embedding space of $f_\theta$ using $(\bm W^*,\bm b^*)$.
When computing the CLIP text embedding for a sample in group $(y,a)$, we do not simply concatenate $t_i^y$ and $t_j^a$ using a single text prompt as in \cite{DRML}. Instead, we average the two text embeddings computed for each $t_i^y$ and $t_j^a$, \textit{i.e., } compute $\frac{1}{2}(\bm z^{\text{CLIP}}_{P_1(t^{y}_i)} + \bm z^{\text{CLIP}}_{P_1(t^{a}_j)})$
as the embedding for $(y,a)$. We empirically find that this way better reflects the individual embeddings of  $t_i^y$ and $t_j^a$. 
Furthermore, to represent the more diverse nature of images with texts, we adopt the 80 CLIP prompt templates used in zero-shot classification in \cite{CLIP}.
That is, when each $(t^{y}_i, t^{a}_j)$ pair is fetched, we compute $\frac{1}{2}(\bm z^{\text{CLIP}}_{P_k(t^{y}_i)} + \bm z^{\text{CLIP}}_{P_k(t^{a}_j)})$, in which $P_k(\cdot)$ is randomly selected among the 80 prompt templates. 


The retraining of the last layer is done by mini-batch optimization as in \cite{SELF}.
Moreover, we sample group-balanced training sets every epoch to maximally utilize available data.
By default, we adopt early stopping based on validation WGA as in \cite{JTT, GDRO, AFR, SELF}. 


%% file: contents/5.experimental_results.tex
\subsection{Dataset Description}
\noindent \textbf{Waterbirds} \cite{GDRO} is a synthetic dataset constructed with CUB \cite{CUB} and Places \cite{Places}.
Minor groups together make up about 5\% of the training dataset.

\noindent \textbf{CelebA} \cite{CelebA} is a large collection of celebrity faces along with 40 attribute annotations per image.
In line with \cite{GDRO}, we consider a classification problem where $\mathcal{Y} = $ \{dark hair, blond hair\} and $\mathcal{A} = $ \{woman, man\}.
The minority group, "blond men," makes up about 1\% of the training dataset.

\noindent \textbf{SpuCoAnimals} \cite{SpuCo} is constructed based on ImageNet-1K \cite{ImageNet} with $\mathcal{Y} = $ \{landbirds, waterbirds, small dogs, large dogs\} and $\mathcal{A} = $ \{land BGs, water BGs, indoor BGs, outdoor BGs\}.
Bird classes are spuriously correlated with land/water BGs and dog classes with indoor/outdoor BGs.
This dataset provides a more realistic setting beyond the binary $\mathcal{Y}, \mathcal{A}$. 
The total proportion of minority groups is about 5\%.

\setlength{\tabcolsep}{4pt}
\begin{table*}[t]
\centering
\caption{
    \label{tab:main_results}
    Test WGA \& average accuracy for each dataset.
    The reported experimental results of each baseline are given in S.M., and we did our best to reproduce the reported results of each method.
    \checkmark \checkmark denotes utilizing group annotated image validation set for training the last layer, on the other hand, a single \checkmark denotes using the set for only hyperparameter searching or model selection.
    $\star$ denotes the setting where ERM model is used, \textit{i.e.} no additional split of training dataset or class-balancing. Refer to \cref{sec:given_erm} for a detailed experimental setting.
    We calculated average accuracy in the same way with \cite{GDRO}.
    All numbers are averaged from 4 random seeds and the highest WGAs are bolded among the last three rows that share the same settings.
}
\vspace{-0.1in}
\input{Table/main_results_table}
\vspace{-0.2in}
\end{table*}

\subsection{Experimental Setup}
\noindent \textbf{[Model Architecture]}
We used ImageNet pre-trained ResNet-50 \cite{ResNet} as the $(f_\theta, h_\phi)$, with the exception of SpuCoAnimals. In the case of SpuCoAnimals, given that it originated from ImageNet, we used a randomly initialized model to prevent any potential information leakage. Additionally, an experiment was conducted with ViT \cite{ViT} based $(f_\theta, h_\phi)$ to verify the applicability of \ours \ to more generalized vision models, as elaborated in Sec. B.2 of the S.M. We employed CLIP ViT-B/32 throughout our experiments.

\noindent \textbf{[Number of Words Generated]}
We generated 200 synonyms for each category name, except for the "large dog" class in SpuCoAnimals where GPT-3.5 cannot generate more than 100 unique breeds due to its limitations.

\noindent \textbf{[Estimation of the $\bm g$]}
The modality gap was estimated from 1K image-text pairs sampled from the validation set of COCO-Caption dataset \cite{CoCoCaption} unless otherwise stated.

\noindent \textbf{[ReLU on the Projected Embedding]}
Since all embeddings of ResNet undergo the ReLU operation, we applied ReLU to each projected embedding $\bm \Pi(\bm z_T^{\text{CLIP}})$ to minimize the discrepancy. 
If $\bm z_I^{f_\theta}$ possesses real values rather than non-negative values, the ReLU operation can be omitted.

\subsection{Main Experimental Results}

\noindent \textbf{[Main results]}
We present a comparison of \ours \ with other baselines in \cref{tab:main_results}.
It is noteworthy that \ours \ exhibits competitive performance with DFR, which explicitly uses a group-balanced image dataset for LLR, on the Waterbirds and CelebA,
Furthermore, \ours \ even demonstrates superior performance on SpuCoAnimals compared to DFR. 
In addition, \ours \ outperforms AFR and SELF across all datasets.
The lower performance of DFR and SELF on SpuCoAnimals may be attributed to the limited data for each group in the group-balanced dataset.
The small number may affect both retraining and the evaluation of the validation WGA since DFR and SELF randomly halve the validation set for WGA evaluation, leading to the suboptimal hyperparameter search.
In contrast, \ours \ avoids this issue by neither splitting the validation set nor using it for LLR.



\noindent \textbf{[Post Hoc Utilization of Baselines]}
\label{sec:given_erm}
As previously mentioned, utilizing \ours \ provides a notable advantage in employing a pre-trained model.
Since the usual development process involves training a model with ERM, identifying weaknesses in the model, and subsequently addressing them, having the capability to leverage an already trained model constitutes a significant practical benefit.
To assess the performance of AFR and SELF under circumstances where additional training of the entire model is not feasible due to computational costs, we evaluate their performances in a post hoc manner.
We used half of the validation set as a reweighting dataset for AFR and excluded class-balanced ERM for SELF.

The results are marked with $\star$ in \cref{tab:main_results}.
While AFR's overall performance improves due to the benefit of training the ERM model with the full training dataset, \ours \ still outperforms.
In contrast, the performance of SELF drops significantly because the model is not trained in a class-balanced manner, which implies that class-balanced ERM is critical to maintaining the performance of SELF, necessitating the additional training of the entire model.
On the other hand, \ours \ can be applied to the pre-trained model without any additional training of the whole model, which is efficient and practical for use.

\noindent \textbf{[Adjustment the Proportion of Minority Examples]}\label{sec:minority_ratio}

We compared \ours \ with other baselines by adjusting the proportion of minority groups from both the training and validation sets.
Unlike the experiments conducted by \cite{DFR, AFR, SELF} where only the proportion of the validation set is controlled, our experimental setting is more realistic; usually training and validation sets are split from an entire dataset, so the proportion changes in both.
We excluded SpuCoAnimals since we observed a case where the validation WGAs are all zero for all hyperparameters as the ratio becomes smaller, making a fair comparison between the baselines challenging.
For a fair comparison, we avoided early stopping for all methods as discussed in \cite{AFR}.

\cref{fig:figure3_1} illustrates that \ours \ consistently outperforms AFR and SELF across all cases and even has superior performance to DFR on Waterbirds.
The performance of all baselines is inevitably influenced by the ratio, considering the diminished size of minority groups in their reweighting dataset.
Moreover, the reduced ratio may impact SELF's ability to identify minority data, as the scarcity of minority groups in the training data might not provide sufficient information to discern differences in predictions between the early-stopped model and the fully trained model.
In contrast, \ours \ has fairly robust performance across varying ratios, making it well-suited for situations where images belonging to minority groups are rare.

\newcommand{\rulesep}{\unskip\ \vrule\ }

\begin{figure*}[t]
  \centering
  \begin{subfigure}[b]{0.28\textwidth}
    \includegraphics[width=\textwidth, height=7.5cm]{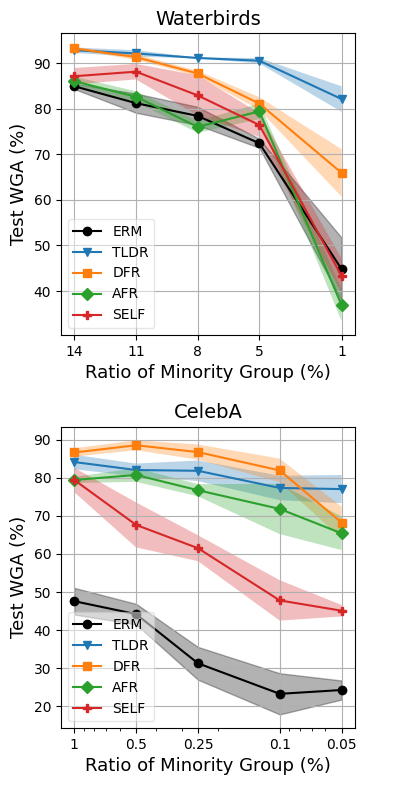}
    \caption{}
    \label{fig:figure3_1}
  \end{subfigure}
  \hfill
  \rulesep
  \begin{subfigure}[b]{0.62\textwidth}
    \includegraphics[width=\textwidth, height=7.5cm]{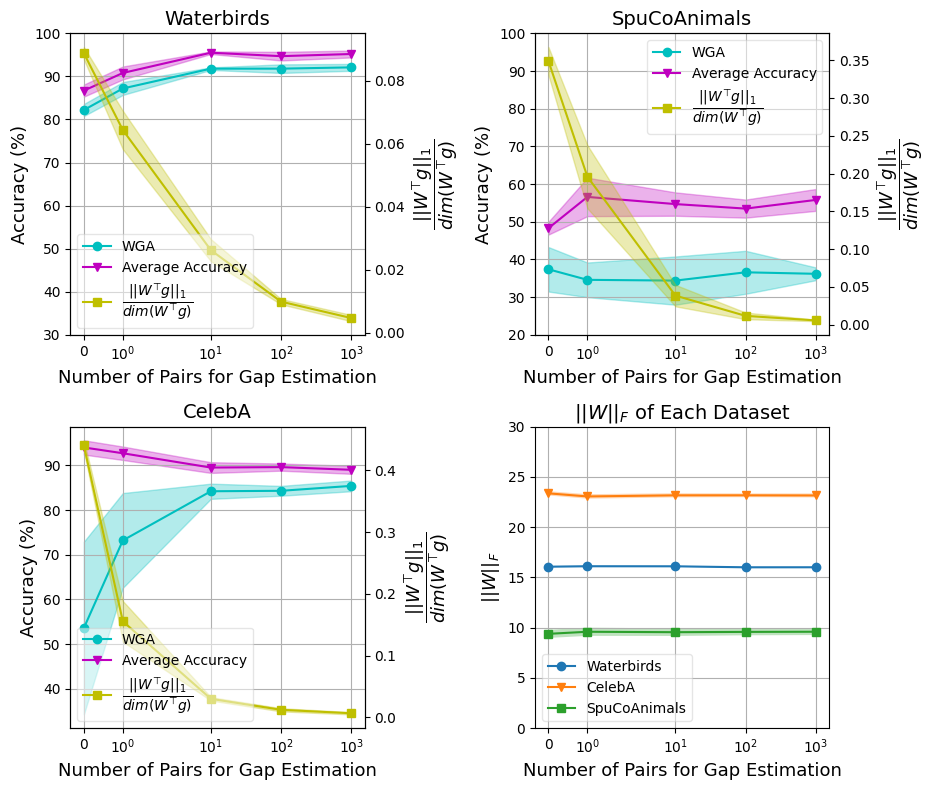}
    \caption{}
    \label{fig:figure3_2}
  \end{subfigure}
  \vspace{-0.1in}
  \caption{
  All numbers are averaged from 4 random seeds.
  \subref{fig:figure3_1}:  Experimental results of controlling the minority ratio on Waterbirds and CelebA.
  \subref{fig:figure3_2}: Experimental results of controlling the number of pairs for gap estimation on all datasets.
  }
  \vspace{-0.2in}
  \label{fig:figures_side_by_side}
\end{figure*}

\subsection{Analyses on the Projector}
\label{sec:pi_analysis}

\noindent \textbf{[Direct Alignment of $\bm \Pi \bm z_T^\text{CLIP}$ and $\bm z_I^{f_\theta}$]} 
One may suggest that direct alignment of $\bm \Pi \bm z_T^\text{CLIP}$ and $\bm z_I^{f_\theta}$ would work using the image-text pairs in COCO-Caption.
However, we experimentally found that approach to be ineffective since it does not consider the dataset on which the $(f_\theta, h_\phi)$ were trained.
This oversight can impact both the VEA as well as the LLR process.
Since the goal is to debias $f_\theta$ within its trained domain, it's crucial for $\bm \Pi$ to perform well within that domain.
Estimating $\bm \Pi$ with extensive $(I,T)$ pairs may inundate it with redundant information, failing to precisely capture the data distribution relevant to ($f_\theta$, $h_\phi$).

We estimated $\bm \Pi_{\text{COCO}}$ by minimizing $||\bm z_I^{f_\theta} - \bm \Pi \bm z_T^{\text{CLIP}}||_2^2$ with the validation split of COCO, 80\% for training ($\# \approx 160$K) and 20\% for validation.
As shown in \cref{tab:options_for_pi} (first row), the aforementioned issue negatively impacts the VEA process for $f_\theta$ and significantly undermines performance on CelebA. 
In addition, \cref{tab:options_for_pi} (second row) demonstrates that using the $\bm \Pi_{\text{COCO}}$ for LLR remains ineffective, even if the VEA is still done with $\bm \Pi_{\text{ours}}$ which is obtained with \cref{lemma_proj} and the datasets used for training $(f_\theta, h_\phi)$.
Hence, estimating $\bm \Pi$ with the same data used for training of ($f_\theta$, $h_\phi$) becomes imperative, thereby validating the novelty and effectiveness of our approach.

\begin{table}[ht]
    \caption{Comparison of the estimation schemes of $\bm \Pi$.}
    \vspace{-0.1in}
    \centering
    \resizebox{\columnwidth}{!}{%
    \begin{tabular}{c c c cc c cc c cc}
    
    \toprule
    \multirow{2}{*}{\textbf{VEA}} & \multirow{2}{*}{\textbf{LLR}} &\multicolumn{2}{c}{\textbf{Waterbirds}} & \multicolumn{2}{c}{ \textbf{CelebA}}
    & \multicolumn{2}{c}{ \textbf{SpuCoAnimals}}
    \\
    
    \cline{3-8}
    \\[-3mm]
    & &  Worst(\%) &  Mean(\%) & 
     Worst(\%) &  Mean(\%)
    &
     Worst(\%) &  Mean(\%)
    
    \\
    
    \midrule

        $\bm \Pi_{\text{COCO}}$ &$\bm \Pi_{\text{COCO}}$ & $86.5_{\pm 0.8}$ & $97.4_{\pm 0.0}$ & $5.6_{\pm 1.2}$ & $90.4_{\pm 1.1}$& $21.0_{\pm 3.2}$ & $70.7_{\pm 2.6}$ \\
        $\bm \Pi_{\text{ours}}$ &$\bm \Pi_{\text{COCO}}$ & $86.8_{\pm 0.3}$ &$97.2_{\pm 0.1}$ &$83.8_{\pm 1.4}$ &$87.3_{\pm 0.4}$ & $26.4_{\pm 5.5}$&$63.0_{\pm 3.1}$\\
        $\bm \Pi_{\text{ours}}$ & $\bm \Pi_{\text{ours}}$ & $\mathbf{92.1}_{\pm 0.3}$&$95.2_{\pm 0.8}$ &$\mathbf{85.4}_{\pm 1.2}$&$89.0_{\pm 0.9}$ & $\mathbf{36.2}_{\pm 1.7}$&$55.8_{\pm 2.9}$\\
    \bottomrule
    \end{tabular}
    }
    \label{tab:options_for_pi}
    \vspace{-0.2in}
\end{table}

\noindent \textbf{[Number of Pairs for Estimation of $\bm g$]} 
We varied the number of pairs used to estimate the modality gap from 0 to 1K to check how our method is affected by the number of pairs.
The results are illustrated in \cref{fig:figure3_2}.
Notably, without gap information, the WGA and average accuracy significantly decrease.
It is noteworthy that a mere 10 image-text pairs suffice to estimate the modality gap, yielding comparable performance to cases with a larger number of pairs.
This suggests a remarkably low burden in terms of pair collection.
However, the experiment result on SpuCoAnimals exhibits a different tendency.
This is because the searched $\lambda$ value is considerably larger compared to the other two datasets, resulting in a smaller $||\bm W||_F$ and,  consequently, reduced $||\bm W^\top \bm g||_1$.
Therefore, the efficacy of the constraint $\bm W^\top \bm g = 0$ is not adequately represented.

\subsection{Analyses on Text Dataset Construction}
\label{sec:drml_compare}
\noindent \textbf{[Comparison with DrML]}
We re-emphasize that DrML \cite{DRML} merely uses words that are already provided in the metadata. 
Therefore, it cannot be seamlessly applied to CelebA, SpuCoAnimals, or other general datasets, which lack appropriate texts in metadata.
In addition, using only words in the metadata may be insufficient due to their limited quantity or misalignment with the embedding spaces, which can lead to poor debiasing performance.
To demonstrate the superiority of our text dataset construction procedure, we carried out experiments on Waterbirds drawing comparison with DrML.
We assessed the performance on the CLIP backbone, comparing DrML's text data against ours, both with and without the inclusion of VEA in \cref{tab:rebut_table3}.
First of all, it is noteworthy that VEA has a positive impact also on the DrML's text data, improving both the test WGA and the average accuracy.
Furthermore, we observe that \ours's text generation scheme \textit{with} VEA is significantly superior to that of DrML, and the VEA step is essential as relying solely on all the generated words proves to be less effective.
We believe that these results clearly show the novelty of \ours's text dataset construction procedure.


\begin{table}[t]
    \caption{
    Comparison between DrML and \ours \ for the construction of text datasets.}
    \vspace{-0.1in}
    \centering
    \resizebox{\columnwidth}{!}{%
    \begin{tabular}{c c c c c c}
        \toprule
        \multirow{2}{*}{\textbf{Method}}& \multirow{2}{*}{\textbf{VEA}} & \textbf{Need} &\textbf{Ratio of}&\multicolumn{2}{c}{\textbf{Waterbirds}}\\

        \cline{5-6}
        & & \textbf{Metadata?} & \textbf{Remaining Words} & Worst(\%) & Mean(\%)
        \\
        \midrule
        CLIP w/ Linear Classifier & - & - & - & $33.5_{\pm 0.6}$ & $96.1_{\pm 0.1}$ \\
        [1mm]\hline
        \multirow{2}{*}{DrML} & \ding{55} & \checkmark & 204 / 204 &$54.8_{\pm 0.3}$ & $93.3_{\pm 0.4}$ \\
         & \checkmark & \checkmark & 181 / 204 &
        $\mathbf{59.5}_{\pm 0.4}$ & $94.2_{\pm 0.2}$
        \\[1mm]\hline

        \multirow{2}{*}{DrML w/ TLDR's Words} & \ding{55} & \ding{55} &800 / 800& $47.1_{\pm0.7}$ & $96.1_{\pm0.0}$ \\
        & \checkmark & \ding{55} & 545 / 800& $\mathbf{73.1}_{\pm1.0}$& $92.3_{\pm{0.2}}$ \\
        
        \bottomrule
    \end{tabular}
    }

    \label{tab:rebut_table3}
    \vspace{-0.1in}
\end{table}

\noindent \textbf{[Ablation study on VEA]}
\label{ablation_filter}
We present the results of ablation studies on the VEA in \cref{tab:ablation_filter}.
Indeed, the test WGA improves on Waterbirds and CelebA.
In addition, it significantly reduces the variance of the test WGA on SpuCoAnimals.
We note that the decrease of the test WGA on SpuCoAnimals is not decisive, considering the substantial variance in the test WGA when VEA is not applied.
The gain from VEA becomes larger when the early stopping is not applied, which can be practical in a situation where the ratio of minority groups is too low to allow proper early stopping.

\begin{table}[t]
\centering
\caption{
\label{tab:ablation_filter}
Result of ablation studies on the VEA.
ES stands for early stopping.
}
\vspace{-0.1in}
\input{Table/filter_ablation_table}
\vspace{-0.2in}
\end{table}

%% file: Table/main_results_table.tex
\resizebox{0.8\textwidth}{!}{
\begin{tabular}{c c c c cc c cc c cc}

\toprule
\multirow{2}{*}{\textbf{Method}} & \textbf{Group Info} & \textbf{Post-hoc}&\multicolumn{2}{c}{\textbf{Waterbirds}} && \multicolumn{2}{c}{\textbf{CelebA}} && \multicolumn{2}{c}{\textbf{SpuCoAnimals}}
\\

\cline{4-5}
\cline{7-8}
\cline{10-11}

\\[-3mm]
& Train / Val & &
Worst(\%) & Mean(\%)
&& Worst(\%) & Mean(\%)
&& Worst(\%) & Mean(\%)
\\







\midrule




ERM &
\ding{55} / \ding{55} & -
& $72.2_{\pm 0.7}$ & $98.1_{\pm1.1}$
&& $47.6_{\pm3.1}$ & $95.2_{\pm0.1}$
&& $6.3_{\pm 1.6}$ & $81.3_{\pm 0.9}$

\\

Group-DRO \cite{GDRO} &
\checkmark / \checkmark & \ding{55}
& $88.2_{\pm0.5}$ & $93.3_{\pm0.7}$ 
&& $90.3_{\pm 0.3}$ & $92.3_{\pm1.9}$ 
&& $39.5_{\pm 4.8}$ & $47.1_{\pm 3.6}$
\\

\DFRVAL \cite{DFR} & 
\ding{55} / \checkmark\checkmark & \checkmark

& $92.5_{\pm0.7}$ & $94.8_{\pm 0.3}$
&& $86.6_{\pm1.1}$ & $90.3_{\pm 0.2}$
&& $22.4_{\pm2.4}$ & $68.4_{\pm 1.1}$
\\[1mm]\hline


AFR \cite{AFR} &
\ding{55} / \checkmark & \ding{55}
&$87.4_{\pm 0.8}$ & $90.4_{\pm0.6}$
&& $79.4_{\pm 0.8}$ & $91.7_{\pm0.3}$
&& $16.2_{\pm6.1}$ & $59.9_{\pm3.5}$

\\

SELF \cite{SELF}&
\ding{55} / \checkmark & \ding{55}
&$91.4_{\pm 2.1}$ & $94.5_{\pm 1.6}$
&& $79.4_{\pm 3.2}$& $91.9_{\pm 0.7}$
&& $7.3_{\pm 2.9}$ & $86.7_{\pm 1.0}$
\\[1mm]\hline

$^\star$AFR &
\ding{55} / \checkmark & \checkmark
& $86.3_{\pm1.9}$ & $91.7_{\pm 0.4}$
&& $80.1_{\pm 3.3}$ & $90.6_{\pm 1.0}$
&& $22.3_{\pm3.4}$ & $53.8_{\pm7.7}$
\\

$^\star$SELF &
\ding{55} / \checkmark & \checkmark
&$91.2_{\pm 0.7}$ & $96.0_{\pm 0.7}$
&& $56.9_{\pm 4.9}$ & $95.0_{\pm 0.1}$
&& $6.9_{\pm0.6}$ & $77.2_{\pm 2.2}$
\\







\ours &
\ding{55} / \checkmark & \checkmark
&$\mathbf{92.1}_{\pm 0.3}$ & $95.2_{\pm 0.8}$
&&$\mathbf{85.4}_{\pm 1.2}$& $89.0_{\pm 0.9}$
&&$\mathbf{36.2}_{\pm1.7}$ & $55.8_{\pm 2.9}$
\\

\bottomrule
\end{tabular}
}

%% file: Table/filter_ablation_table.tex
\resizebox{\columnwidth}{!}{%
\begin{tabular}{c c c cc c cc c cc}

\toprule
\multirow{2}{*}{\textbf{ES}} & \multirow{2}{*}{\textbf{VEA}} &\multicolumn{2}{c}{\textbf{  Waterbirds}} & \multicolumn{2}{c}{ \textbf{CelebA}}
& \multicolumn{2}{c}{ \textbf{SpuCoAnimals}}
\\

\cline{3-8}
\\[-3mm]
& &  Worst(\%) &  Mean(\%) & 
 Worst(\%) &  Mean(\%)
&
 Worst(\%) &  Mean(\%)

\\

\midrule
\checkmark & \ding{55}
&  $91.1_{\pm0.8}$ &   $93.1_{\pm1.3}$
&  $84.1_{\pm1.2}$ &  $87.8_{\pm1.5}$
&  $\mathbf{38.9}_{\pm7.0}$ &  $51.3_{\pm2.5}$

\\



\checkmark & \checkmark
&  $\mathbf{92.1}_{\pm0.3}$ &  $95.2_{\pm0.8}$
&  $\mathbf{85.4}_{\pm1.2}$ &  $89.0_{\pm0.9}$
&  $36.2_{\pm1.7}$ &  $55.8_{\pm2.9}$ \\ \hline

\ding{55} & \ding{55}
&  $88.1_{\pm0.7}$ &  $88.6_{\pm0.5}$
&  $76.0_{\pm2.5}$ &  $82.2_{\pm1.2}$
&  $\mathbf{34.7}_{\pm7.7}$ &  $49.9_{\pm3.3}$ \\
\ding{55} & \checkmark
&  $\mathbf{92.0}_{\pm0.4}$ &  $95.6_{\pm0.2}$
&  $\mathbf{84.1}_{\pm1.9}$ &  $88.0_{\pm1.8}$
&  $28.3_{\pm2.9}$ &  $52.1_{\pm1.5}$ \\

\bottomrule
\end{tabular}%
}












%% file: contents/6.conclusion.tex
In this study, we demonstrate that a general image classifier can be debiased with text-based LLR.
\ours \ stands out by not necessitating a group-annotated, balanced image dataset, nor requires additional training of the entire model, making it easily applicable.
In addition, TLDR introduces a systematic procedure for constructing text datasets that leverages debiasing performance.
Nevertheless, there are some limitations to our method.
As our work is based on CLIP, we can leverage the concepts that CLIP understands.
We anticipate that this limitation can be addressed with the advancements in multi-modal contrastive models across various domains, such as ConVIRT \cite{ConVIRT} for the medical domain and WikiSatNet \cite{WikiSatNet} for satellite imagery.
Additionally, our approach assumes prior knowledge of the model's weakness.
Considering the plenty of works focused on discovering the model's weaknesses \cite{Domino, SEAL, Spotlight, GEORGE}, we anticipate that acquiring knowledge about the model's weaknesses can be readily achieved.
Lastly, there may be concerns about the inherent bias of LLMs and CLIP \cite{LLMBias, CLIPBias}, but we did not encounter any substantial negative impact on our results.

%% file: contents/supplementary.tex
\definecolor{codegreen}{rgb}{0,0.6,0}
\definecolor{codegray}{rgb}{0.5,0.5,0.5}
\definecolor{codepurple}{rgb}{0.58,0,0.82}
\definecolor{backcolour}{rgb}{0.95,0.95,0.92}

\lstdefinestyle{pythonstyle}{
  backgroundcolor=\color{backcolour}, commentstyle=\color{codegreen},
  keywordstyle=\color{magenta},
  numberstyle=\tiny\color{codegray},
  stringstyle=\color{codepurple},
  basicstyle=\ttfamily\footnotesize,
  breakatwhitespace=false,         
  breaklines=true,                 
  captionpos=b,                    
  keepspaces=true,                 
  numbersep=5pt,                  
  showspaces=false,                
  showstringspaces=false,
  showtabs=false,                  
  tabsize=2
}
\lstset{
basicstyle=\small\ttfamily,
columns=flexible,
breaklines=true
}

%


\section{Additional Related Works}
\subsection{Prior Works on Mitigating Spurious Correlation}
\label{sec:supp_relwork}
Plenty of works have been suggested to mitigate spurious correlation in classification and these can be categorized according to the assumption of the accessibility to group annotations and knowledge of spurious correlation.

\subsubsection{With Fully Available Group Annotations}
In a context where group annotations for all data are fully available, Group-DRO  \cite{GDRO} proposed an online optimization algorithm that reduces the loss of the worst-performing group.
In addition, \cite{SUBG} demonstrated that straightforward group balancing of the training dataset is effective for mitigating spurious correlation without introducing any additional hyperparameters.
These works are often regarded as the maximum achievable performance due to completely available group annotations.
Nevertheless, the acquisition of the group annotations of the entire dataset requires human labor, which introduces huge costs.



\subsubsection{With Group Annotations of the Validation Set}
Recognizing the difficulties in obtaining group annotations for the entire dataset, various approaches have been suggested to improve the accuracy of the minority group by exploiting group annotations from the validation set only.
SSA \cite{SSA} adopts a semi-supervised approach, employing group-annotated validation data to train a group label predictor, subsequently creating pseudo-group annotations for the training data.
Then, they utilize Group-DRO \cite{GDRO} with these pseudo-group annotations to achieve group robustness.
DFR \cite{DFR} has experimentally demonstrated that even if a model is biased towards spurious attributes, the feature extractor can still adequately learn the core features.
They argue that the satisfactory worst group accuracy can be achieved through last-layer retraining with a group-balanced validation set.
However, these methods still have the limitation of requiring a group-annotated image validation set for training.
In addition, DFR necessitates a group-balanced image validation set which can limit its applicability.

\subsubsection{Without Group Annotations and Knowledge on Spurious Correlation}
Under circumstances where group annotations as well as knowledge of the type of spurious correlation cannot be obtained, methods for inferring which data belongs to minority groups have been introduced \cite{JTT, LfF, CnC, AFR, SELF}.
LfF \cite{LfF} trains two neural networks simultaneously; one intentionally biased and the other debiased.
Concurrently, the \textit{debiased} network is trained to focus on samples that the biased model finds challenging. 
This is done by reweighting the training samples based on their relative difficulty determined by the cross entropy loss of both models.
JTT \cite{JTT} initially trains a reference model for a few epochs, and then examples misclassified by this reference model are identified to be belonging to minority groups.
They subsequently upsample these misclassified examples and train a new model using the upsampled dataset. 
These methods have a significant drawback: they involve numerous hyperparameters which makes hyperparameter tuning time-consuming and their performance is highly sensitive to these hyperparameters.
CnC \cite{CnC} adopts a contrastive learning approach to learn representations that are robust to spurious correlations.
Different from previous methods, CnC utilizes the outputs of a trained ERM model to identify samples within the same class but possessing dissimilar spurious features.
Our baselines, AFR \cite{AFR} and SELF \cite{SELF} also fall into this category as they do not require group annotated image dataset for training, nor prior knowledge of spurious correlations present in the dataset.
Hence, one can employ these methods in situations where knowledge of the model's vulnerabilities is lacking.
Nevertheless, most of the methods require time-consuming hyperparameter tuning and they still have subpar performances compared to DFR or \ours.
In addition, most methods require additional training of the entire model or a secondary model, which makes them less practical.



\subsection{Related Works on Using Texts for Vision Models}
\input{contents/2.related_works}
\section{Additional Experimental Results}
\subsection{Reported Results of Baselines with ResNet-50}

\begin{table}[t]
\centering
\caption {
    \label{tab:reported_values}
    Reported test WGA \& average accuracy of each baseline with ResNet-50 backbone.
}
\input{Table/reported_values}
\end{table}

We present the reported results of each baseline in \cref{tab:reported_values} for the reader's information, which are omitted in \cref{tab:main_results} of the manuscript.

\subsection{Main Results with Vision Transformer}
\label{sec:main_results_vit}

\begin{table}[t]
\centering
\caption{
    \label{tab:main_results_vit}
    Test WGA \& average accuracy for each dataset with ViT-B/16 backbone.
    All numbers are averaged from 4 random seeds and the highest WGAs are bolded among the last three rows that share the same settings.
}
\input{Table/main_results_vit_table}
\end{table}

To show that our method can be applied to more general architecture, we conducted an experiment with a vision transformer.
We used ViT-B/16 of which pre-trained weight is provided from \texttt{timm} package.
We used the same hyperparameter search space with \cref{tab:main_results} of the manuscript except for slight modification.
Please refer to \cref{sec:exp_details_vit} for detailed descriptions of hyperparameters.
In addition, we did not apply ReLU to the projected embedding $\bm \Pi(z_T^{\text{CLIP}})$ as an embedding from ViT-B/16 possesses real values.
The result is summarized in \cref{tab:main_results_vit}.
Notably, \ours \ is effective for mitigating spurious correlation in a vision transformer-based architecture which validates that \ours \ can be applied to the general architecture.

\subsection{Additional Ablation Studies}
\input{contents/Supp_additional_ablation_studies}

\section{Additional Analyses}
\subsection{Effect of Orthogonality} 
To validate that orthogonality between $\bm W$ and $\bm g$ is essential to achieve cross-modal transferability within the embedding space of a general image classifier, we employed the COCO-Caption dataset \cite{CoCoCaption} where explicit image-text pairs exist.
We randomly sampled $2 \times 5000$ image-text pairs from the dataset to construct the training and validation sets.
We calculated the $(\bm W^*, \bm b^*)$ of $\bm \Pi$ with/without the constraint $\bm W^\top \bm g = 0$ as outlined in \cref{lemma_proj} of the manuscript using the training set, and evaluated the degree of orthogonality ($\frac{||\bm W^\top \bm g||_1}{\dim(\bm W^\top \bm g)}$) and the proximity between the projected text embedding and the corresponding image embedding of $f_\theta$ ($\frac{||\bm z_{I}^{f_\theta} - \bm \Pi(\bm z_{T}^{\text{CLIP}})||_1}{\dim(\bm z_{I}^{f_\theta})}$) with the validation set.
We set $\lambda = 0$ to isolate the impact of the constraint, as altering the value of $\lambda$ can influence both the norm of $\bm W$ and the proximity between embeddings.
The results are summarized in \cref{tab:gap_reg_coco}.
The findings affirm that ensuring orthogonality between $\bm W$ and $\bm g$ contributes to bringing the projected text embedding closer to its corresponding image embedding in the embedding space of $f_\theta$.

\begin{table}[t]
\centering
\caption{
\label{tab:gap_reg_coco}
Effect of orthogonality on cross-modal transferability.
}
\input{Table/gap_reg_ablation_table}

\end{table}

\subsection{Modality Gap Without $\ell_{2}$ Normalization}
\label{sec:ablation_normalization}

\begin{table}[t]
    \centering
    \caption{Average magnitude and direction of each $\bm z_{I_i}^{\text{CLIP}} - \bm z_{T_i}^{\text{CLIP}}$ when normalization is applied or not.}
    \input{Table/constant_gap_table}

\label{tab:ablation_normalization}
\end{table}

\begin{figure}[t]
  \centering
  \begin{subfigure}[b]{0.45\columnwidth}
    \includegraphics[width=\columnwidth, height=5cm]{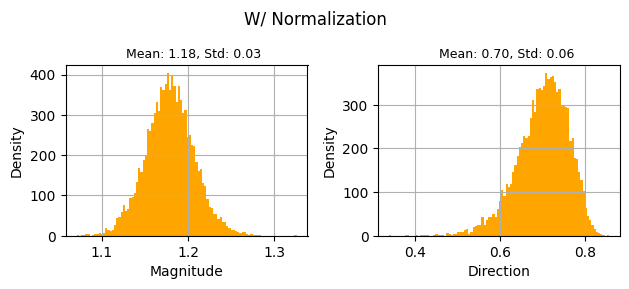}
    \caption{}
    \label{fig:w_normalization}
  \end{subfigure}
  \begin{subfigure}[b]{0.45\columnwidth}
    \includegraphics[width=\columnwidth, height=5cm]{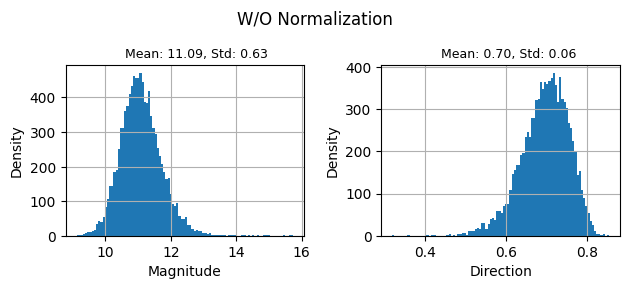}
    \caption{}
    \label{fig:wo_normalization}
  \end{subfigure}
  \hfill
  \caption{
  \subref{fig:w_normalization} : Histogram of magnitudes and directions of each $\bm z_{I_i}^{\text{CLIP}} - \bm z_{T_i}^{\text{CLIP}}$ with $\ell_2$ normalization of each $\bm z_{I_i}^{\text{CLIP}}, \bm z_{T_i}^{\text{CLIP}}$.
  \subref{fig:wo_normalization} : Histogram of magnitudes and directions of each $\bm z_{I_i}^{\text{CLIP}} - \bm z_{T_i}^{\text{CLIP}}$ without $\ell_2$ normalization of each $\bm z_{I_i}^{\text{CLIP}}, \bm z_{T_i}^{\text{CLIP}}$.
  }
  \label{fig:normalization}
\end{figure}
As stated in \cref{sec:proj} of the manuscript, we do not normalize each CLIP embedding as usually done.
This is because normalization of embeddings can degrade the performance of alignment between two embedding spaces due to computational precision as discussed in \cite{TextToConcept}.
In addition, we found empirically that the averaging of embeddings mentioned in \cref{sec:llr} of the manuscript does not work effectively for normalized embeddings.
We defer the details on this to \cref{sec:ablation_umap}.

We first demonstrate that the modality gap is nearly constant despite the absence of $\ell_{2}$ normalization of CLIP embeddings.
We sampled 10K image-text pairs from COCO-Caption dataset \cite{CoCoCaption} and observed the distribution of magnitudes $||\bm z_{I_i}^{\text{CLIP}} - \bm z_{T_i}^{\text{CLIP}}||$ and directions $\cos (\bm z_{I_i}^{\text{CLIP}} - \bm z_{T_i}^{\text{CLIP}}, \bm \hat g)$ of each gap following \cite{DRML}.

The results are shown in \cref{tab:ablation_normalization} and \cref{fig:normalization}.
It is noticeable that the gap of each image-text pair is almost constant even though each CLIP embedding is not $\ell_2$ normalized, implying that the assumption of constant modality gap is valid.


\begin{figure}[t!]
  \centering
  \begin{subfigure}[b]{0.9\columnwidth}
    \includegraphics[width=\columnwidth, height=5cm]{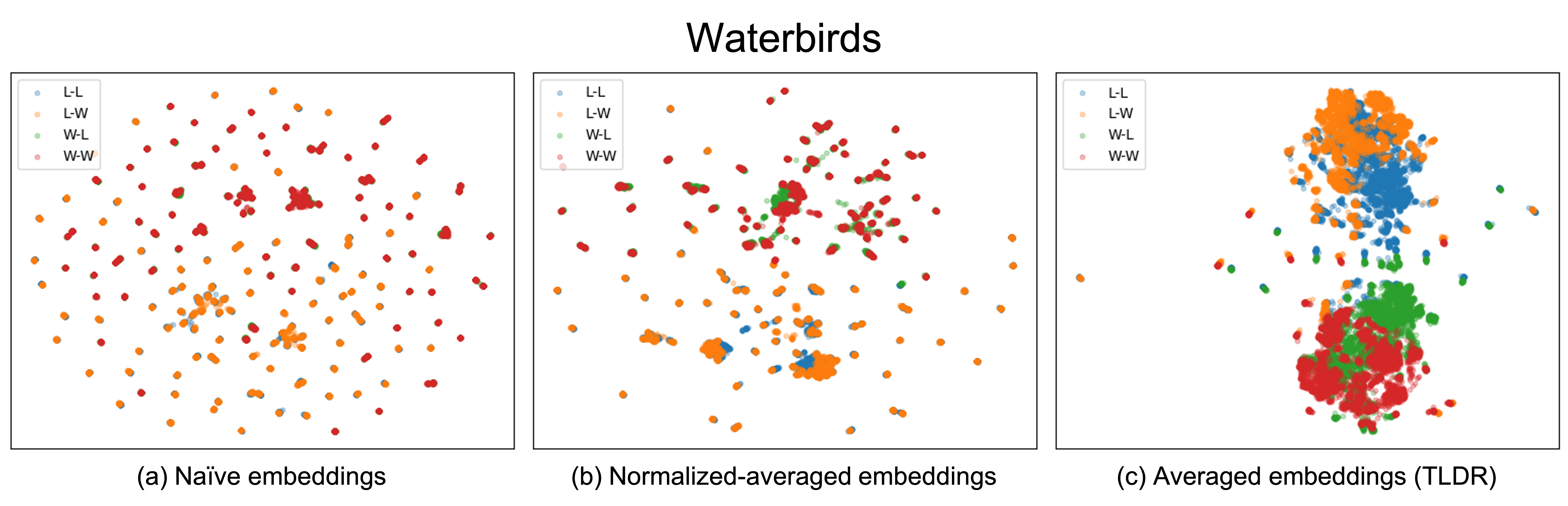}

    \label{fig:wb_embedding}
  \end{subfigure}

  \begin{subfigure}[b]{0.9\columnwidth}
    \includegraphics[width=\columnwidth, height=5cm]{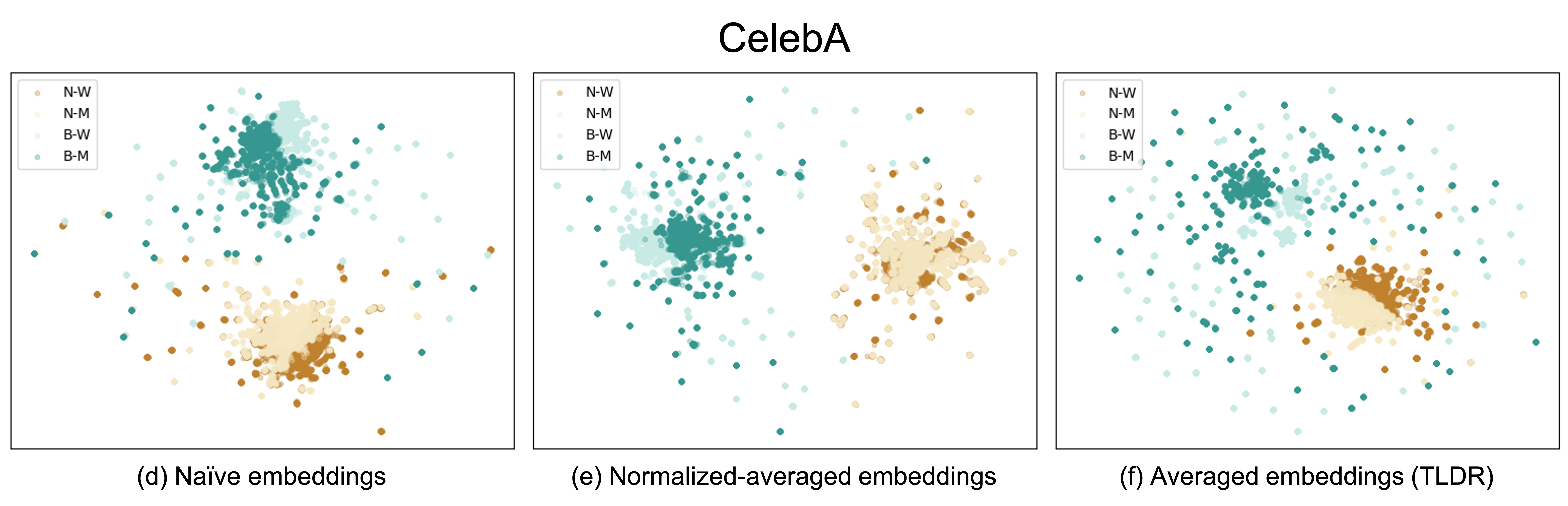}
    
    \label{fig:celeba_embedding}
  \end{subfigure}
 \begin{subfigure}[b]{0.9\columnwidth}
    \includegraphics[width=\columnwidth, height=5cm]{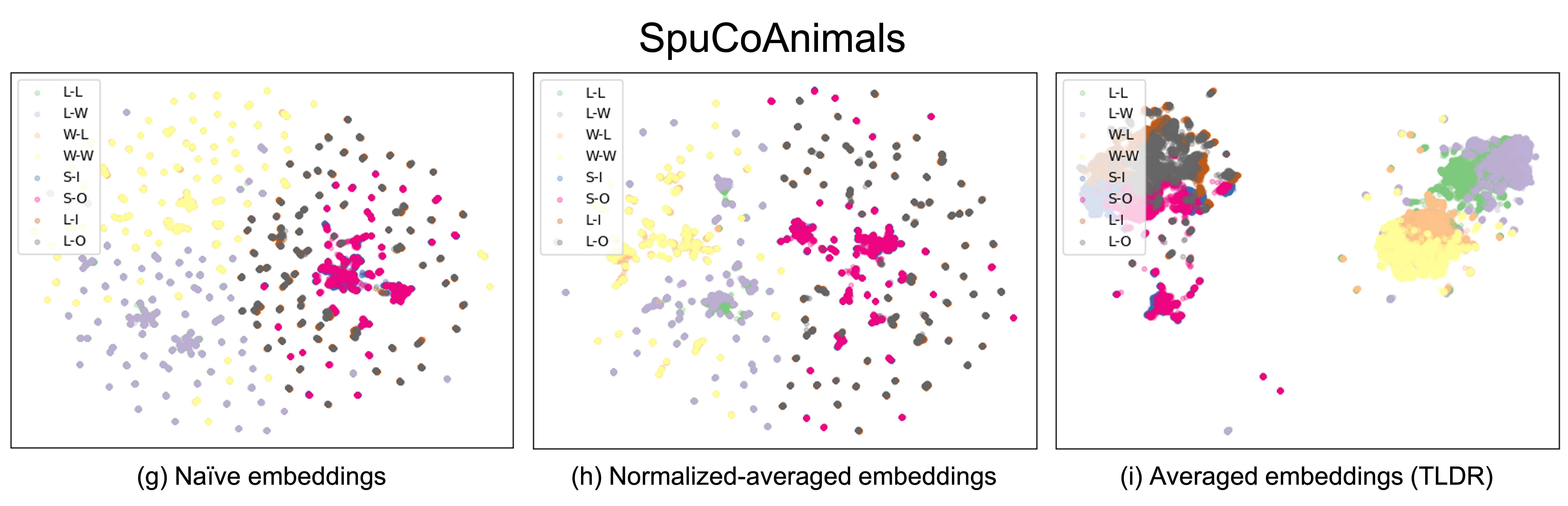}
    \label{fig:spuco_embedding}
  \end{subfigure}
  \caption{
    Figure of UMAP projected CLIP text embeddings of each dataset. 
    We randomly sampled 5000 pairs of $(t^y_i, t^a_j)$ for each group for clear visualization.
    We abbreviate groups of each dataset as follows.
    Waterbirds: \{(L)andbirds / (W)aterbirds - (L)and backgrounds / (W)ater backgrounds\}, 
    CelebA: \{(N)on blond / (B)lond - (W)omen / (M)en\},
    SpuCoAnimals: \{(L)andbirds / (W)aterbirds - (L)and backgrounds / (W)ater backgrounds, (S)mall dogs / (L)arge dogs - (I)ndoor backgrounds / (O)utdoor backgrounds\}.
    }

  \label{fig:umap}
\end{figure}

\subsection{UMAP Based Analysis on Averaged Embeddings}
\label{sec:ablation_umap}

As explained in \cref{sec:llr} of the manuscript, we use averaged embeddings, \textit{i.e., } $\frac{1}{2} (\bm z^{\text{CLIP}}_{P_1(t^{y}_i)} + \bm z^{\text{CLIP}}_{P_1(t^{a}_j)})$ for a clear separation between groups, and we refer to these embeddings as \textit{averaged embeddings} in this section.
To illustrate, consider prompts ``A photo of a girl." and ``A photo of golden hair.".
We computed the embeddings of each prompt and then take their average.
This approach contrasts with what we call \textit{naive embeddings}, utilized in DrML \cite{DRML}.
An example of a \textit{naive embedding} is the embedding of the prompt ``A photo of a girl with golden hair.".
The list of prompt templates for \textit{naive embeddings} of each dataset is as follows.
\begin{itemize}
    \item Waterbirds : \lstinline|"A photo of a {|$t^y_i$\lstinline|} in the {|$t^a_j$\lstinline|}."|
    \item CelebA : \lstinline|"A photo of a {|$t^a_j$\lstinline|} with {|$t^y_i$\lstinline|}."|
    \item SpuCoAnimals : \lstinline|"A photo of a {|$t^y_i$\lstinline|} in the {|$t^a_j$\lstinline|}."|
    
\end{itemize}

In \cref{fig:umap}, we illustrate UMAP \cite{UMAP} projected embeddings residing in the CLIP embedding space.
It is noticeable that \textit{naive embeddings} (\cref{fig:umap} (a), (d), (g)) exhibit overlap between groups, especially groups that share $t^y_i$.
This implies that the presence of $t^a_j$ has only a marginal effect on the separation between groups, suggesting that the CLIP embedding space puts more emphasis on $t^y_i$.
In contrast, \textit{averaged embeddings} (\cref{fig:umap} (c), (f), (i)) provide a better distinction between groups compared to \textit{naive embeddings}, suggesting that \textit{averaged embeddings} better capture the diversity and unique characteristics of each group.

In addition, as stated in \cref{sec:ablation_normalization}, we tried averaging the two embeddings which are both $\ell_2$ normalized, which is referred to as \textit{normalized-averaged embeddings} in this section.
That is, we used $\frac{1}{2}(\tilde{\bm z}^{\text{CLIP}}_{P_1(t^{y}_i)} + \tilde{\bm z}^{\text{CLIP}}_{P_1(t^{a}_j)})$ where $\tilde{\bm z} = \frac{\bm z}{||\bm z||_2}$.
From \cref{fig:umap} (b), (e), and (h), it can be noticed that averaging after normalization of embedding does not separate between groups effectively.
This is one of the reasons why the CLIP embeddings are not normalized in our work.
Consequently, we opt for averaging unnormalized embeddings.

\input{contents/Supp_discussions}

\section{Proof of \cref{lemma_proj}}
\label{sec:proof_proj}
\input{contents/Supp_proof_of_lemma}

\section{Experimental Details}
\label{sec:ExpDetails}

\input{contents/Supp_experimental_details}

\subsection{Dataset Configuration}
\label{sec:dataset_config}

\begin{table}[t]
\caption{Configurations of each dataset.}

\input{Table/data_configuration}
\label{tab: dataset}
\end{table}
\begin{table}[ht!]
\centering
\caption{Configuration of Waterbirds in \cref{fig:figure3_1}.}
\input{Table/Nominority_waterbirds_configuration}
\label{tab: no_minor_waterbirds}
\end{table}

We summarize configurations of each dataset in \cref{tab: dataset}.
All of the datasets have imbalanced data distributions, with a very low proportion of minority groups.
Especially, Waterbirds has a distribution shift between training and validation sets, which is unusual given that training and validation sets are typically split from a single dataset.
Hence, we combine the training and validation sets, then randomly split them in an 8:2 ratio in \cref{fig:figure3_1} of the manuscript.
The newly split Waterbirds are illustrated in \cref{tab: no_minor_waterbirds}.

\subsection{Full List of Prompt Templates}
\label{sec:templates}
\lstset{style=pythonstyle}
\input{contents/Supp_templates}

\noindent \textbf{List of Prompt Templates Used in \cref{sec:dfr_with_syn}}
\label{sec:diffusiontemplates}
\lstset{style=pythonstyle}
\input{contents/Supp_prompt}

%% file: contents/2.related_works.tex
Recently, there has been a noticeable trend towards exploiting texts for vision models for various purposes leveraging information in the image-text joint embedding space generated by ALIGN \cite{jia2021scaling} or CLIP \cite{CLIP}.
It has been applied in data augmentation \cite{TextManiA}, domain generalization \cite{PromptStyler, GroundingVisualRep}, concept-based explanation \cite{TextToConcept, CounTEX}, error slice discovery \cite {Domino} and model selection \cite{LOVM}. 
However, no studies have yet been carried out on the use of text for the debiasing of general image classifiers.
Moreover, prior works mainly project information from the vision model to the joint embedding space to use information from texts \cite{CounTEX, TextToConcept, GroundingVisualRep, Domino} or utilize cross-modal transferability only in the joint embedding space \cite{PromptStyler, LOVM, DRML}.
In contrast, our work focuses on preserving cross-modal transferability in the embedding space of the general image classifier, shedding light on the enjoyment of language-only debiasing for arbitrary vision models.

%% file: Table/reported_values.tex
\resizebox{0.9\textwidth}{!}{
\begin{tabular}{c c c c cc c cc c cc}

\toprule
\multirow{2}{*}{\textbf{Method}} & \textbf{Group Info} & \textbf{Post-hoc}&\multicolumn{2}{c}{\textbf{Waterbirds}} && \multicolumn{2}{c}{\textbf{CelebA}} && \multicolumn{2}{c}{\textbf{SpuCoAnimals}}
\\

\cline{4-5}
\cline{7-8}
\cline{10-11}

\\[-3mm]
& Train / Val & &
Worst(\%) & Mean(\%)
&& Worst(\%) & Mean(\%)
&& Worst(\%) & Mean(\%)
\\
\midrule
Group-DRO&
\checkmark / \checkmark & \ding{55}
& $91.4_{\pm1.1}$ & $93.5_{\pm0.3}$
&& $88.9_{\pm2.3}$ & $92.9_{\pm0.2}$
&& - & -
\\

\DFRVAL& 
\ding{55} / \checkmark\checkmark & \checkmark
& $92.9_{\pm0.2}$ & $94.2_{\pm0.4}$
&& $88.3_{\pm1.1}$ & $91.3_{\pm0.3}$
&& - & -
\\

AFR&
\ding{55} / \checkmark & \ding{55}
& $90.4_{\pm1.1}$ & $94.2_{\pm1.2}$
&& $82.0_{\pm0.5}$ & $91.3_{\pm0.3}$
&& - & -
\\

SELF&
\ding{55} / \checkmark & \ding{55}
& $92.0_{\pm1.3}$ & $94.0_{\pm1.7} $
&& $82.2_{\pm2.8}$ & $91.7_{\pm0.4} $
&& - & -
\\
\bottomrule
\end{tabular}
}

%% file: Table/main_results_vit_table.tex
\resizebox{0.9\columnwidth}{!}{
\begin{tabular}{c c c c cc c cc c cc}
\toprule
\multirow{2}{*}{\textbf{Method}} & \textbf{Group Info} & \textbf{Post-hoc}&\multicolumn{2}{c}{\textbf{Waterbirds}} && \multicolumn{2}{c}{\textbf{CelebA}} && \multicolumn{2}{c}{\textbf{SpuCoAnimals}}
\\

\cline{4-5}
\cline{7-8}
\cline{10-11}

\\[-3mm]
& Train / Val & &
Worst(\%) & Mean(\%)
&& Worst(\%) & Mean(\%)
&& Worst(\%) & Mean(\%)

\\[1mm]\hline\\[-3mm]

ERM &
\ding{55} / \ding{55} & -
& $75.0_{\pm 0.9}$ & $98.5_{\pm0.1}$
&& $48.5_{\pm2.2}$ & $95.5_{\pm0.1}$
&& $7.1_{\pm 1.2}$ & $75.7_{\pm1.0}$

\\

Group-DRO &
\checkmark / \checkmark & \ding{55}
& $89.9_{\pm0.9}$ & $97.5_{\pm0.9}$ 
&& $90.2_{\pm 1.3}$ & $93.5_{\pm0.4}$ 
&& $29.6_{\pm 7.1}$ & $46.4_{\pm 5.3}$
\\

\DFRVAL & 
\ding{55} / \checkmark\checkmark & \checkmark
& $90.1_{\pm0.4}$ & $96.9_{\pm 0.3}$
&& $72.3_{\pm3.5}$ & $79.3_{\pm 1.1}$
&& $36.5_{\pm2.5}$ & $45.4_{\pm 2.2}$

\\[1mm]\hline


AFR &
\ding{55} / \checkmark & \ding{55}
&$81.9_{\pm 4.7}$ & $94.6_{\pm3.4}$
&& $85.4_{\pm 1.1}$ & $91.7_{\pm0.3}$
&& $21.5_{\pm6.1}$ & $49.6_{\pm10.8}$
\\

SELF &
\ding{55} / \checkmark & \ding{55}
&$87.5_{\pm 1.0}$ & $97.6_{\pm 0.2}$
&& $75.8_{\pm 3.5}$& $92.9_{\pm 0.3}$
&& $5.9_{\pm 0.7}$ & $73.5_{\pm 0.5}$
\\[1mm]\hline

$^\star$AFR &
\ding{55} / \checkmark & \checkmark
& $\mathbf{91.1}_{\pm0.6}$ & $95.6_{\pm 0.6}$
&& $80.1_{\pm 1.9}$ & $92.2_{\pm 0.4}$
&& $16.0_{\pm8.2}$ & $56.3_{\pm5.9}$
\\

$^\star$SELF &
\ding{55} / \checkmark & \checkmark
&$87.3_{\pm 1.5}$ & $97.6_{\pm 0.3}$
&& $52.9_{\pm 7.6}$ & $95.2_{\pm 0.3}$
&& $7.7_{\pm2.1}$ & $76.2_{\pm 0.5}$
\\

\ours &
\ding{55} / \checkmark & \checkmark
&$90.0_{\pm 1.2}$ & $92.2_{\pm 1.1}$
&&$\mathbf{81.8}_{\pm 3.9}$& $88.6_{\pm 0.1}$
&&$\mathbf{24.7}_{\pm4.3}$ & $46.8_{\pm 4.0}$


\\[1mm]\hline

\end{tabular}
}

%% file: contents/Supp_additional_ablation_studies.tex
\subsubsection{Effect of Diverse Prompt Templates}
\begin{table}[t]
\centering
\caption{Result of ablation study on diverse prompt templates.}
\input{Table/text_aug_table}    
\label{tab:text_augment}
\end{table}

We conducted an ablation study on utilizing zero-shot classification templates for retraining the last linear layer and the result is shown in \cref{tab:text_augment}.
It can be verified that utilizing diverse prompts is effective for improving overall performance.

\subsubsection{Ablation on Number of Words Generated}
\begin{table}[t]
    \centering
    \caption{Result of ablation study on the number of words generated.}
    \input{Table/number_of_words_ablation}
    \label{tab:number_of_words}
\end{table}

We conducted an ablation study on the number of words generated.
We varied the size of $\mathcal{T}^y, \mathcal{T}^a$ as \{50, 100, 150, 200\} by sampling from the full list of generated words.
The results are shown in \cref{tab:number_of_words}.
The number of words does indeed affect the performance of \ours.
Nevertheless, only 100 words for each category are sufficient to achieve competitive performance when 200 words per category are used.

\subsubsection{Gap Estimation with Another Dataset}
To demonstrate that the modality gap $\bm g$ can be estimated with any dataset including image-text pairs, we estimated the $\bm g$ by sampling 1000 pairs from SBU dataset  \cite{SBU}.
The results in the \cref{tab:gap_dataset} show that $\hat{\bm g}$ is indeed dataset agnostic. 

\begin{table}[t]
    \centering
    \caption{Result of debiasing based on $\bm g$ estimated with SBU.}
    \input{Table/sbu_estimation}
    \label{tab:gap_dataset}
\end{table}

%% file: Table/text_aug_table.tex
\resizebox{0.5\columnwidth}{!}{%
\begin{tabular}{c c cc c cc c cc c cc}
\toprule
\multirow{2}{*}{\textbf{Datasets}} & \multicolumn{2}{c}{\textbf{Only $P_1$}} & \multicolumn{2}{c}{\textbf{Use $P_1, ... , P_{80}$}}
\\

\cline{2-5}
\\[-3mm]
& Worst(\%) & Mean(\%) & 
Worst(\%) & Mean(\%)
\\[1mm]
\midrule

Waterbirds &
$91.9_{\pm 0.5}$ & $93.3_{\pm 0.7}$ & $92.1_{\pm 0.5}$ & $95.4_{\pm 0.5}$ 
\\
CelebA &
$83.2_{\pm 1.2}$ & $89.7_{\pm 0.8}$ & $85.4_{\pm 1.2}$ & $89.0_{\pm 0.9}$

\\
SpuCoAnimals&
$35.1_{\pm 3.6}$ & $57.5_{\pm4.6}$ & $36.2_{\pm 1.7}$ & $55.8_{\pm 2.9}$

\\
\bottomrule
\end{tabular}
}

%% file: Table/number_of_words_ablation.tex
\resizebox{0.75\columnwidth}{!}{%
\begin{tabular}{c c cc c cc c cc c cc c cc}
\toprule
\multirow{2}{*}{\textbf{\# of Words per Category}} & \multicolumn{2}{c}{\textbf{Waterbirds}} & \multicolumn{2}{c}{\textbf{CelebA}} & \multicolumn{2}{c}{\textbf{SpuCoAnimals}}
\\

\cline{2-7}
\\[-3mm]
& Worst(\%) & Mean(\%) & 
Worst(\%) & Mean(\%) &
Worst(\%) & Mean(\%) &
\\[1mm]
\midrule

50 &
$88.7_{\pm 0.7}$ & $93.3_{\pm 0.9}$ & $83.9_{\pm 1.8}$ & $89.5_{\pm 0.5}$ & $34.9_{\pm 10.2}$ & $60.5_{\pm 3.8}$
\\
100 &
$90.6_{\pm 0.6}$ & $94.4_{\pm 1.1}$ & $84.2_{\pm 0.6}$ & $89.3_{\pm 1.3}$ & $36.0_{\pm 2.9}$ & $58.1_{\pm 4.7}$
\\
150 (100 for `large dogs')&
$91.9_{\pm 0.7}$ & $94.9_{\pm 0.7}$ & $84.2_{\pm 1.3}$ & $88.3_{\pm 1.1}$ & $37.1_{\pm 5.6}$ & $55.6_{\pm 2.7}$ 
\\
200 (100 for `large dogs')&
$92.1_{\pm 0.5}$ & $95.4_{\pm 0.5}$ & $85.4_{\pm 1.2}$ & $89.0_{\pm 0.9}$ & $36.2_{\pm 1.7}$ & $55.8_{\pm 2.9}$
\\
\bottomrule
\end{tabular}
}

%% file: Table/sbu_estimation.tex
  \resizebox{0.85\columnwidth}{!}{%
    \begin{tabular}{c c cc c cc c cc c cc c cc}
    \toprule
    \multirow{2}{*}{\textbf{Dataset Used for Gap Estimation}} & \multicolumn{2}{c}{\textbf{Waterbirds}} & \multicolumn{2}{c}{\textbf{CelebA}} & \multicolumn{2}{c}{\textbf{SpuCoAnimals}}
    \\
    
    \cline{2-7}
    \\[-3mm]
    & Worst(\%) & Mean(\%) & 
    Worst(\%) & Mean(\%) &
    Worst(\%) & Mean(\%) &
    \\[1mm]
    
    \midrule
      COCO-Val &$92.1_{\pm0.5}$ & $95.4_{\pm 0.5}$ & $85.4_{\pm 1.2}$ & $89.0_{\pm 0.9}$& $36.2_{\pm1.7} $&$ 55.8_{\pm 2.9}$\\
    SBU Caption &$91.7_{\pm 0.2}$&$93.7_{\pm 0.5}$ & $85.3_{\pm 1.2}$& $88.7_{\pm 1.1}$& $38.7_{\pm 4.7}$&$54.5_{\pm 4.4}$\\
    \bottomrule
    \end{tabular}
    }

%% file: Table/gap_reg_ablation_table.tex
\resizebox{0.45\columnwidth}{!}{
\begin{tabular}{c c c}
\toprule
\textbf{Include $\bm W^\top \bm g = 0$} & $\frac{||\bm W^\top \bm g||_1}{\dim(\bm W^\top \bm g)}$ & $\frac{||\bm z_{I}^{f_\theta} - \bm \Pi(\bm z_{T}^{\text{CLIP}})||_1}{\dim(\bm z_{I}^{f_\theta})}$
\\
\midrule
\ding{55}
&$1.25_{\pm 0.48}$ & $0.87_{\pm 0.44}$
\\
\checkmark
& $0.88_{\pm 0.61}$ & $0.56_{\pm 0.37}$
\\
\bottomrule
\end{tabular}
}

%% file: Table/constant_gap_table.tex
\resizebox{0.35\columnwidth}{!}{%
\begin{tabular}{c|cc}
\toprule
$\ell_2$ \textbf{Normalization} & \textbf{Magnitude}    & \textbf{Direction}    \\
\midrule
Yes & $1.18_{\pm 0.03}$ & $0.70_{\pm 0.06}$ \\
No & $11.09_{\pm 0.64}$ & $0.70_{\pm 0.06}$ \\
\bottomrule
\end{tabular}%
}

%% file: contents/Supp_discussions.tex
\subsection{Validity of Assumption 1}
Since our primary task is classification, it is may be sufficient that $\bm \Pi \bm z_I^{\text{CLIP}}$ is closer to its corresponding $\bm z_I^{f_\theta}$ than the average distance between $\bm z_{I_1}^{\text{CLIP}}, \bm z_{I_2}^{\text{CLIP}}$ where $I_1, I_2$ are belong to the same class with $I$.

To validate the Assumption 1, we compared the intra class NMSE and  NMSE$(\bm \Pi \bm z_I^{\text{CLIP}}, \bm z_I^{f_\theta})$ in \cref{tab:nmse} (row 1-2).
The intra class NMSE denotes the average distance between two arbitrary embeddings $\bm z_{I_1}^{\text{CLIP}}, \bm z_{I_2}^{\text{CLIP}}$ belonging to the same class. We computed the value by sampling 100 embeddings within each class and averaging the NMSE values for all possible pairs, as done in \cite{CounTEX}.
All NMSE values were measured based on the validation split of each benchmark dataset.
As shown in \cref{tab:nmse} (row 1-2), the $\bm \Pi$ effectively maps each $\bm z_I^{\text{CLIP}}$ with significantly lower NMSE values compared to the average intra class variation, supporting the validity of our assumption. 

\subsection{Analysis on the Effect of Variation in Modality Gap}
As the modality gap is not exactly constant, there is a possibility that the efficacy of the constraint $\bm W^\top \hat {\bm g} = 0$ may be called into question.
We argue the effect of the variability in the modality gap is minimal in the sense that the value of NMSE$(\bm \Pi \bm z_T^{\text{CLIP}}, \bm z_I^{f_\theta})$ remains lower than the intra class NMSE.
However, the measurement of NMSE$(\bm \Pi \bm z_T^{\text{CLIP}}, \bm z_I^{f_\theta})$ requires image-text pairs, which are not included in the benchmark datasets. In addition, the more accurate analysis necessitates the paired image of each $\bm z_T^{\text{CLIP}}$ used for LLR, which are hard to collect.
Nevertheless, we measured the value with the validation split of Waterbirds by generating captions using the metadata and prompt templates $P_1, ..., P_{80}$. While the value may not be identical to that obtained with the texts used for LLR, we believe it to be sufficiently similar.
As demonstrated in  \cref{tab:nmse} (row 3), the NMSE$(\bm \Pi \bm z_T^{\text{CLIP}}, \bm z_I^{f_\theta})$ remains lower than the intra class NMSE. This indicates that the variation in the modality gap may influence the projection of $\bm z_T^{\text{CLIP}}$, but not to a considerable extent.



\begin{table}[t]
    \centering
    \caption{NMSE for all possible pairs within each class for all datasets}
    \resizebox{0.65\columnwidth}{!}{%
    \begin{tabular}{l c c c}
    \toprule
     & \textbf{Waterbirds} & \textbf{CelebA} &\textbf{SpuCoAnimals}
    \\[1mm]
    \midrule
    
        Intra class NMSE & $0.5931_{\pm 0.0140}$ & $0.4851_{\pm 0.0261}$& $0.5056_{\pm 0.0839}$ \\
        NMSE$(\bm \Pi \bm z_I^{\text{CLIP}}, \bm z_I^{f_\theta})$ &$0.2109_{\pm 0.0013}$ &$0.1896_{\pm 0.0019}$ & $0.1135_{\pm 0.0079}$ \\
        NMSE$(\bm \Pi \bm z_T^{\text{CLIP}}, \bm z_I^{f_\theta})$, $\bm W^\top \hat {\bm g} = 0$ & $0.3481_{\pm 0.0008}$ & -& -\\
        NMSE$(\bm \Pi \bm z_T^{\text{CLIP}}, \bm z_I^{f_\theta})$, Adding $\hat {\bm g}$ & $0.4275_{\pm 0.0030}$ & - & - \\
    \bottomrule
    \end{tabular}
    }

    \label{tab:nmse}
\end{table}

\subsection{Analysis on Modality Gap Mitigation Approaches}
We compared the other possible approaches to mitigate the modality gap issue; adding Gaussian noise when training $\bm \Pi$ \cite{Icantbelieve} or projecting $\bm z_I^{\text{CLIP}}$ into CLIP's text embedding \cite{decap}.

For the case of adding Gaussian noise, given that the $\bm \Pi$ is estimated using $\bm z_I^{\text{CLIP}}$, we added the noise to $\bm z_I^{\text{CLIP}}$ to reduce the modality gap.
The experiment was conducted in two cases: one in which the $\bm \Pi$ was estimated using ridge regression, and the other in which it was optimized using stochastic gradient descent (SGD) to minimize the mean squared error (MSE) loss, $MSE(\bm \Pi \bm z_I^{\text{CLIP}}, \bm z_I^{f_\theta})$.
Furthermore, the variance of the Gaussian noise was tuned.
On the other hand, for the projection of $\bm z_I^{\text{CLIP}}$ to CLIP's text embeddings, it was necessary to choose the support text set.
We considered two options; a set of captions included in the training split of COCO Caption, as selected by \cite{decap}, and a set constructed based on our generated words and prompt templates $P_1, ..., P_{80}$.
The experimental results are reported in \cref{tab:modality_gap_mitigation}.
It can be checked the other approaches are not as effective as ours.

Adding Gaussian noise has limitations in that adding noise to  $\bm z_I^{\text{CLIP}}$ can result in erratic outcomes as highlighted by \cite{Icantbelieve}. In addition, it requires cumbersome hyperparameter searching to tune the variance of the Gaussian noise.
On the other hand, a key challenge in projecting $\bm z_I^{\text{CLIP}}$ into text embeddings, suggested by \cite{decap}, lies in the choice of an appropriate text support set, since an arbitrary choice has been shown to be ineffective (see \cref{tab:modality_gap_mitigation} row 3); when the COCO-train captions are used as the support set, the training fails.
In light of these observations, we assert that our method offers a clear advantage over other approaches in simplicity and effectiveness.

Furthermore, one may suggest that recovering $\bm z_I^{\text{CLIP}}$ corresponding to $\bm z_T^{\text{CLIP}}$ by adding $\hat {\bm g}$ to each $\bm z_T^{\text{CLIP}}$ from the relation $\bm g_i = \bm z_{I_i}^{\text{CLIP}} - \bm z_{T_i}^{\text{CLIP}}$.
Note that the variation in the modality gap results in an error of $\bm W^\top(\hat {\bm g} - \bm g_i)$ on $\bm \Pi$ since $\bm \Pi(\bm z_{T_i}^{\text{CLIP}} + \hat {\bm g}) = \bm \Pi(\bm z_{T_i}^{\text{CLIP}} + \bm g_i - \bm g_i + \hat {\bm g}$ = $\bm \Pi(\bm z_{I_i}^{\text{CLIP}} - \bm g_i + \hat {\bm g}) \approx \bm z_{I_i}^{f_\theta} + \bm W^\top(\hat {\bm g} - \bm g_i)$ where $\bm g_i = \bm z_{I_i}^{\text{CLIP}} - \bm z_{T_i}^{\text{CLIP}}$.
For the sake of argument, let us assume that $\bm g_i$ has a similar direction to $\hat {\bm g}$ but differs in magnitude, i.e., $\bm g_i \approx c \hat {\bm g}$ for some constant $c$.
In this case, the error $\bm W^\top(\hat {\bm g} - \bm g_i) \approx \bm W^\top ((1-c) \hat {\bm g})$ is not equal to 0 without the orthogonality constraint.

The error introduced by the variation in the modality gap is still equal to $\bm W^\top(\hat {\bm g} - \bm g_i)$ in our approach because $\bm \Pi(\bm z_{T_i}^{\text{CLIP}}) = \bm \Pi(\bm z_{I_i}^{\text{CLIP}} - \bm g_i) \approx \bm z_{I_i}^{f_\theta} - \bm W^\top (\bm g_i) = \bm z_{I_i}^{f_\theta} + \bm W^\top (\hat{\bm g} - \bm g_i)$.
On the contrary, our approach does not introduce any error as the constraint guarantees that $\bm W^\top(\hat {\bm g} - \bm g_i) \approx \bm W^\top ((1-c) \hat {\bm g}) = 0$. 
We believe that this difference contributed to the result of larger NMSE$(\bm \Pi \bm z_T^{\text{CLIP}}, \bm z_I^{f_\theta})$ with the simple addition of $\hat {\bm g}$ compared to that of our approach. (See \cref{tab:nmse} row 3-4.)
In addition, we believe that the larger NMSE values hurt the performance of LLR, which can be checked from \cref{tab:modality_gap_mitigation} (row 5).
From these, we posit that the simple addition of $\hat {\bm g}$ is not as effective as our approach.

\begin{table}
    \centering
    \caption{Comparison of modality gap mitigation approaches.}

  \resizebox{0.8\columnwidth}{!}{%
    \begin{tabular}{l c cc c cc c cc c cc c cc}
    \toprule
    \multirow{2}{*}{} & \multicolumn{2}{c}{\textbf{Waterbirds}} & \multicolumn{2}{c}{\textbf{CelebA}} & \multicolumn{2}{c}{\textbf{SpuCoAnimals}}
    \\
    
    \cline{2-7}
    \\[-3mm]
    & Worst(\%) & Mean(\%) & 
    Worst(\%) & Mean(\%) &
    Worst(\%) & Mean(\%) &
    \\[1mm]
    
    \midrule
        Gaussian + Ridge &$89.5_{\pm2.6 }$ & $92.5_{\pm 2.7}$ & $82.5_{\pm 0.9}$ & $84.9_{\pm0.6 }$ & $\mathbf{38.7}_{\pm 5.3}$& $50.9_{\pm 1.5}$ \\
        Gaussian + SGD w/ MSE Loss &$91.5_{\pm 0.8}$ & $94.2_{\pm 1.1}$&$82.4_{\pm 1.7}$ & $85.0_{\pm 1.0}$ & $33.0_{\pm4.8}$  & $52.7_{\pm 4.5}$\\
        Projection w/ COCO train &$77.1_{\pm 3.1}$ & $96.4_{\pm 0.2} $ & $2.9_{\pm0.8}$ & $89.1_{\pm0.8}$& $4.6_{\pm4.3}$ & $61.9_{\pm9.6}$ \\
        Projection w/ our generated words & $90.9_{\pm0.6}$ & $91.8_{\pm 0.6}$ & $79.4_{\pm 1.9}$ & $89.8_{\pm 0.7}$& $31.4_{\pm4.4}$ &  $63.5_{\pm 3.7 }$ \\
        Adding $\bm \hat{g}$& $91.1_{\pm0.2}$ & $95.2_{\pm0.5}$ & $84.2_{\pm 1.1}$ & $89.3_{\pm1.6}$& $35.6_{\pm 9.6}$ & $56.7_{\pm 3.1}$ \\
        \ours & $\mathbf{92.1}_{\pm 0.3}$ & $95.2_{\pm 0.8}$
        &$\mathbf{85.4}_{\pm 1.2}$& $89.0_{\pm 0.9}$
        &$36.2_{\pm1.7}$ & $55.8_{\pm 2.9}$
        \\
    \bottomrule
    \end{tabular}
    }
    \label{tab:modality_gap_mitigation}
\end{table}
\subsection{Computational Time}
The computation time for computing $X^\top X$ and $(X^\top X)^{-1}$ is reported for each dataset in \cref{tab:computation_time}. Each reported time is the average of 10 runs and was measured on a single RTX A5000 GPU. Additionally, the computation was performed using the \texttt{torch.linalg.inv} function.
The results demonstrate that the computation burden is insignificant.

\begin{table}[t]
    \centering
    \caption{Computation time for $X^\top X$ and $(X^\top X + \lambda I)^{-1}$}
    \resizebox{0.55\columnwidth}{!}{%
    \begin{tabular}{c c c c}
    \toprule
     & \textbf{Waterbirds} & \textbf{CelebA} &\textbf{SpuCoAnimals}
    \\[1mm]
    \midrule
    
        $X^\top X$ (ms) & $0.078_{\pm 0.001}$ & $0.135_{\pm 0.003}$  & $0.125_{\pm 0.003}$\\
        $(X^\top X + \lambda I)^{-1}$ (ms) &$10.581_{\pm 0.505}$ & $209.857_{\pm 0.082}$& $56.897_{\pm 0.006}$\\
    \bottomrule
    \end{tabular}
    }
    \label{tab:computation_time}
\end{table}

\subsection{Tradeoff between WGA and Mean Accuracy}
It should be noted that the reported mean accuracy is the weighted average accuracy where the weights are defined as the frequency of each group in the training split, firstly introduced by \cite{GDRO}.
As the benchmark datasets are highly skewed, which can be checked from \cref{tab: dataset}, the weighted average accuracy can be dominated by the performance of majority groups.
Therefore, it can drop as the performance of the majority groups declines while that of the worst groups improves.
Nevertheless, since the primary goal of debiasing is to enhance the WGA, we argue that the decline in the mean accuracy is not as considerable as the enhancement in the WGA.
Moreover, we present the unweighted accuracy in \cref{tab:conventional_acc} for comparison. It can be observed that the unweighted accuracy of TLDR exceeds that of ERM.

\begin{table}[t]
    \centering
    \caption{Unweighted accuracy of ERM and TLDR.}

    \resizebox{0.55\columnwidth}{!}{%
    \begin{tabular}{c c c c}
    \toprule
     & \textbf{Waterbirds} & \textbf{CelebA} &\textbf{SpuCoAnimals}
    \\[1mm]
    \midrule
    
        ERM unweighted Acc. & $91.6_{\pm 0.3} $  & $75.4_{\pm 1.7}$& $50.4_{\pm 0.5}$ \\
      TLDR unweighted Acc. & $93.9_{\pm 0.4}$ & $88.8_{\pm 0.7}$ & $51.8_{\pm 1.3}$ \\
    \bottomrule
    \end{tabular}
    }
    \label{tab:conventional_acc}
\end{table}

\subsection{DFR with Synthetic Images}
\label{sec:dfr_with_syn}

With the advancement of image generation models, several works have investigated utilizing diffusion models to extend real datasets for domain adaptation \cite{yuan2022not}, semi-supervised learning \cite{you2023diffusion}, or generating data augmentations \cite{bansal2023leaving, trabucco2023effective}. Recently, \cite{qraitem2023fake} have introduced the use of synthetic data for bias mitigation. 
To investigate the efficacy of using \textit{synthetic images} instead of \textit{texts} in LLR, we conducted an experimental study.
The dataset is created based on the Stable-Diffusion v1.5, employing 1-2 prompts per group within each dataset. 
These prompts are detailed in \cref{sec:diffusiontemplates}. 
When generating synthetic images using prompts, the guidance scale $\alpha = 2, 4, 8$ was adjusted to regulate the diversity of the generated images. A smaller alpha results in a more diverse range of images, while a larger alpha produces more consistent images closely related to the prompts.
Subsequently, we generated 200 images per group for a fair comparison with \ours \ then DFR is performed with these images. 

\cref{tab:synthetic} shows the results where Syn ($\alpha = 2, 4, 8$) indicates the dataset which only consists of synthetic images with the guidance scale $\alpha$.
It is notable that DFR with synthetic images shows a lower test WGA than ERM's.
While there is some evidence of mitigation of bias on CelebA and SpuCoAnimals, these results are still inferior to those obtained with \DFRVAL or \ours.

\begin{figure*}
\includegraphics[width=\textwidth]{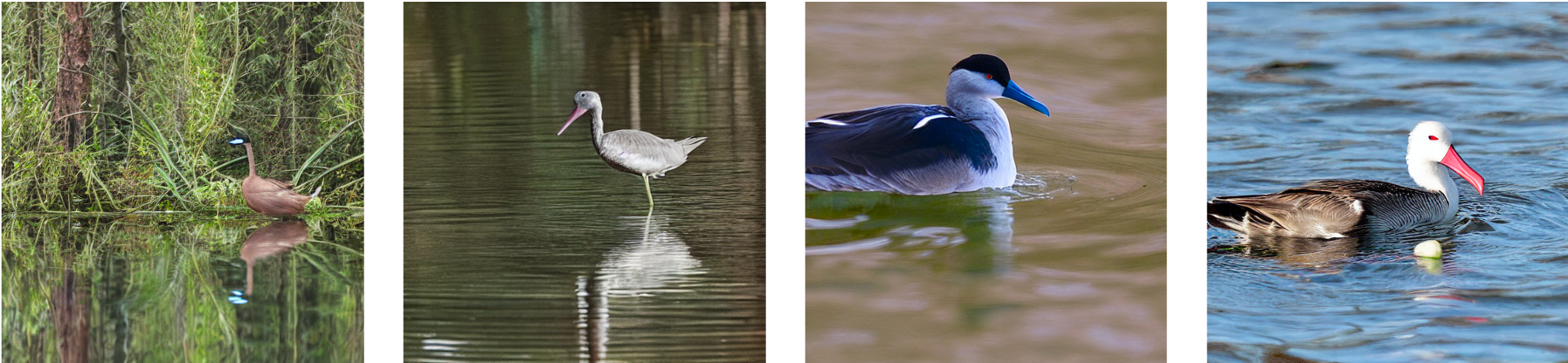} 
\vspace{-0.2in}
\caption{ 
The images depicted above represent instances belonging to the (waterbirds, land background) group. Many of these instances inaccurately feature ‘water’ backgrounds instead of ‘land’ ones. 
}
\label{fig:synthetic_false_example}
\end{figure*}

We attribute the inferior debiasing results of DFR based on synthetic images to \textit{distribution shift} and \textit{inherent bias} of the diffusion models.
First, there may be a covariate shift between the original training data and the generated images.
For example, the Waterbirds dataset consists of composite images based on CUB and Places, so the images contain unrealistic parts or artifacts, while the generated images do not.
We emphasize that this domain mismatch is not unique to Waterbirds, so it is necessary to collect the group-balanced dataset of which data distribution is well matched to that of the datasets on which the ERM model is trained.
In contrast, \ours \ reflects the data distribution on which the ERM model is trained by training the projector $\bm \Pi$.
Thanks to $\bm \Pi$, the text embedding of CLIP can be well adapted to the distribution without raising the issue of covariate shift.
In addition, there is an inherent bias in the text-to-image generation model itself.
From \cref{fig:synthetic_false_example}, it can be observed that the diffusion model fails to correctly generate the images corresponding to \texttt{"A photo of a waterbird in the land background"}.
We suggest that this failure is due to the bias inherent in the diffusion model, which correlates \textit{waterbirds} with \textit{water backgrounds}.
In contrast, \ours \ effectively gets around this problem by explicitly adding text embeddings of $\mathcal{A}$ to those of $\mathcal{Y}$ in the CLIP embedding space.
For these two reasons, we posit that the naive use of synthetic images for DFR does not effectively mitigate the bias of the ERM model.




\setlength{\tabcolsep}{4pt}
\begin{table*}[t]

\centering
\caption{
    \label{tab:synthetic}
    Test WGA \& average accuracy for each dataset with synthetic images.
    DFR$^\mathcal{D'}_\mathcal{D}$ indicates that the ERM model is trained with $\mathcal{D}$ and the DFR is performed with $\mathcal{D'}$.
}
\input{Table/synthetic}

\end{table*}

%% file: Table/synthetic.tex












\resizebox{0.65\columnwidth}{!}{
\begin{tabular}{c c c c c c c c c}
\toprule
\multirow{2}{*}{\textbf{Method}} & \multicolumn{2}{c}{\textbf{Waterbirds}} && \multicolumn{2}{c}{\textbf{CelebA}} && \multicolumn{2}{c}{\textbf{SpuCoAnimals}} \\
\cline{2-3} \cline{5-6} \cline{8-9}
& Worst(\%) & Mean(\%) && Worst(\%) & Mean(\%) && Worst(\%) & Mean(\%) \\
\midrule
ERM & $72.2_{\pm 0.7}$ & $98.1_{\pm 1.1}$ && $47.6_{\pm 3.5}$ & $95.2_{\pm 0.1}$ && $6.3_{\pm 1.6}$ & $81.3_{\pm 0.9}$ \\
DFR$_\text{Tr}^{\text{Syn}(\alpha=2)}$ & $71.7_{\pm 3.9}$ & $87.3_{\pm 1.2}$ && $67.1_{\pm 6.9}$ & $86.3_{\pm 6.1}$ && $21.2_{\pm 4.6}$ & $54.6_{\pm 0.8}$ \\
DFR$_\text{Tr}^{\text{Syn}(\alpha=4)}$ & $66.1_{\pm 3.9}$ & $94.9_{\pm 0.3}$ && $73.3_{\pm 9.2}$ & $80.2_{\pm 8.5}$ && $13.8_{\pm 0.9}$ & $55.6_{\pm 0.8}$ \\
DFR$_\text{Tr}^{\text{Syn}(\alpha=8)}$ & $68.1_{\pm 1.8}$ & $95.1_{\pm 0.4}$ && $79.4_{\pm 5.8}$ & $84.4_{\pm 5.4}$ && $13.9_{\pm 4.7}$ & $52.9_{\pm 2.0}$ \\
\DFRVAL & $92.5_{\pm 0.7}$ & $94.8_{\pm 0.3}$ && $86.6_{\pm 1.1}$ & $90.3_{\pm 0.2}$ && $22.4_{\pm 2.4}$ & $68.4_{\pm 1.1}$ \\
\ours & $\mathbf{92.1_{\pm 0.3}}$ & $95.2_{\pm 0.8}$ && $\mathbf{85.4_{\pm 1.2}}$ & $89.0_{\pm 0.9}$ && $\mathbf{36.2_{\pm 1.7}}$ & $55.8_{\pm 2.9}$ \\
\bottomrule
\end{tabular}
}

%% file: contents/Supp_proof_of_lemma.tex
\begin{proof}
We extend the Lemma 2.1.1. in \cite{MVLR} by adding $\ell_2$-regularization term.

Considering $d_{f_\theta} = 1$ case, the optimization problem is reduced to 
   $\min ||X\bm W - Y||^2_2 + \lambda||\bm W||_2^2$ with $\bm W^\top \bm g = 0$.
Note that the $\bm W$ is a column vector as $d_{f_\theta} = 1$.
Let Lagrangian of this problem as $\mathcal{L}(\bm W; \bm \nu) = ||X\bm W - Y||^2_2 + \lambda||\bm W||^2_2 + \nu(\bm W^\top \bm g)$.
Then, we can get $\bm W^*$ by solving equation $\left.\frac{\partial \mathcal{L}}{\partial \bm W} = 0\right\vert_{\bm W^*, \nu^*}$ where $\nu^*$ is solution of dual problem.

\begin{align}
\left.\frac{\partial \mathcal{L}}{\partial \bm W}\right\vert_{\bm W^*, \nu^*} &= 2X^\top X\bm W^* - 2X^\top Y + 2\lambda \bm W^* + \nu^* \bm g = 0 \\
&\Leftrightarrow \bm W^* = (X^\top X+\lambda I)^{-1}(X^\top Y - \frac{1}{2}\nu^*\bm g)
\label{eq:pf_1}
\end{align}
Plug-in the $\bm {W}^*$ into the constraint $\bm W^\top \bm g = 0$.
\begin{align}
    &\bm W^* \bm g = 0 \\
    &\Leftrightarrow ((X^\top X + \lambda I)^{-1}X^\top Y)^\top \bm g = \frac{\nu^*}{2}((X^TX+\lambda I)^{-1}\bm g)^\top \bm g \\
    &\Leftrightarrow \nu^* = 2(\bm g^\top (X^\top X+\lambda I)^{-1}\bm g)^{-1}\bm g^T \tilde{\bm W}
\end{align}
where $\tilde{\bm W} = (X^\top X + \lambda I)^{-1}X^\top Y$.\\
Then, plug-in the $\nu^*$ into Equation \ref{eq:pf_1}.
\begin{align}
    \bm W^* &= \tilde{\bm W} - \frac{1}{2}(X^\top X + \lambda I)^{-1}\bm g \nu^* \\
    &= \tilde{\bm W} - (X^\top X + \lambda I)^{-1}\bm g(\bm g^\top (X^\top X+\lambda I)^{-1}\bm g)^{-1}\bm g^T \tilde{\bm W} \\
\end{align}
Also, it is obvious that $\bm b^* = \frac{1}{n} (Y-X\bm W^*)^\top \mathbbm{1}.$

As in the proof of Lemma 2.1.1. in \cite{MVLR}, we can generalize this to where $d_{f_\theta} > 1$, then we get the $\bm W^*, \bm b^*$ as in the statement.
\end{proof}

%% file: contents/Supp_experimental_details.tex
\noindent \textbf{Codes}
Our code is constructed on SpuCo\footnote{\url{https://github.com/bigml-cs-ucla/spuco}}, and reproduced AFR\footnote{\url{https://github.com/AndPotap/afr}} and SELF\footnote{\url{https://github.com/tmlabonte/last-layer-retraining}} based on their released codes.

\noindent \textbf{Augmentation of each dataset}
\begin{itemize}
    \item \textbf{Waterbirds}: We used random crops 
    (\texttt{RandomResizedCrop(224, scale=(0.7, 1.0), ratio=(0.75, 4/3), interpolation=2)}) and horizontal flips \\ (\texttt{RandomHorizontalFlip(p=0.5)}) provided from \texttt{torchvision.transforms}.
    \item \textbf{CelebA}: We used random crops (\texttt{RandomResizedCrop(224, scale=(0.7, 1.0), ratio=(1, 4/3), interpolation=2)}) and horizontal flips \\ (\texttt{RandomHorizontalFlip(p=0.5)}) provided from \texttt{torchvision.transforms}.
    \item \textbf{SpuCoAnimals} : We did not use any data augmentation following \cite{SpuCo}.
\end{itemize}

\subsection{Details of VEA with CLIP on SpuCoAnimals}
As "water BG" and "land BG" have some similarities with "outdoor BG", we apply the semantic filter separately for each spurious attribute.
That is, we check whether 
\begin{align}
\argmax_{a'_i \in \mathcal{A}, 1 \leq i \leq 2}  \texttt{cosine-similarity}(\bm z^{\text{CLIP}}_{P_1(t^{a}_i)}, \bm z^{\text{CLIP}}_{P_1(a'_i)}) = a
\end{align}
for each generated word for ``water BG" and ``land BG" where $a'_i$ denotes $i$-th element of $\mathcal{A}$.
On the other hand, we check whether 
\begin{align}
\argmax_{a'_i \in \mathcal{A}, 3 \leq i \leq 4}  \texttt{cosine-similarity}(\bm z^{\text{CLIP}}_{P_1(t^{a}_i)}, \bm z^{\text{CLIP}}_{P_1(a'_i)}) = a
\end{align}
for each generated word for ``indoor BG" and ``outdoor BG".
For consistency, we apply the semantic filter to $\mathcal{T}^y$ in the same way.

\subsection{Number of Remaining Words After VEA}

We summarize the number of remaining words after VEA in \cref{tab:rebut_table2}.
\begin{table}[t]
    \centering
    \caption{Number of words after VEA}

    \resizebox{0.8\columnwidth}{!}{%
    \begin{tabular}{c|c|c|c|c|c}
        \toprule
         & Waterbirds & CelebA & SpuCoAnimals & DrML's Words in \cref{tab:rebut_table3} & TLDR's Words in \cref{tab:rebut_table3} \\
        \midrule
      $|\mathcal{T}^{y}|$ &[96, 125] & [169, 69] & [95, 148, 147, 81] & [135, 42] & [95, 151] \\
      $|\mathcal{T}^{a}|$ &[169, 130] & [120, 189] & [167, 135, 130, 188] & [2, 2] & [169, 130]\\
      \bottomrule
    \end{tabular}
    }
    \label{tab:rebut_table2}
\end{table}

\subsection{Details of Hyperparameter Search}
\label{sec:hyper_search}
\subsubsection{Waterbirds}
\begin{itemize}
    \item \textbf{ERM}:
    We used SGD as the optimizer with a batch size of 32 and trained the model for 300 epochs without any scheduler.
    We searched learning rate from \{1e-4, 1e-3, 3e-3, 1e-2\} and weight decay from \{1e-4, 1e-3, 1e-2\}.
    It is used for DFR and TLDR while SELF and AFR have their own ERM stage.

    {
    \item \textbf{Group-DRO}:
    We used SGD as the optimizer with a batch size of 128 and trained the model for 300 epochs without any scheduler.
    We searched learning rate and weight decay from a set of pairs \{(1e-5, 1.0), (1e-4, 1e-1), (1e-3, 1e-4)\} and $\eta_q$ (learning rate for weights of each group) from \{1e-4, 1e-3, 1e-2, 1e-1\}.
    }
    \item \textbf{DFR}:
            \begin{itemize}
                \item ERM stage: We used the same hyperparameter configuration with the aforementioned ERM model.
                \item LLR stage: We searched $\ell_1$ penalty from \{1e-2, 3e-2, 7e-2, 1e-1, 3e-1, 7e-1, 1.0\}.
            \end{itemize}
            
    \item \textbf{AFR}:
            \begin{itemize}
                \item ERM stage: We used SGD as the optimizer with a batch size of 32 and trained the model for 50 epochs with a cosine annealing scheduler.
                We searched learning rate from \{1e-4, 1e-3, 3e-3, 1e-2\} and weight decay from \{1e-4, 1e-3, 1e-2\}.
                \item LLR stage: We trained the model for 500 epochs.
                We searched $\gamma$ (specifies how much to upweight examples with poor predictions) from 13 points linearly spaced between $[4, 10]$, learning rate from \{1e-2, 2e-2, 3e-2\} and $\lambda$ (specifies how much to keep the original weight) from \{0, 1e-1, 2e-1, 3e-1, 4e-1\}.
            \end{itemize}
    \item \textbf{SELF}:
            \begin{itemize}
                \item Class-balanced ERM stage: We used SGD as the optimizer with a batch size of 32 and trained the model for 100 epochs with a cosine annealing scheduler.
                We searched learning rate from \{1e-4, 1e-3, 3e-3, 1e-2\} and weight decay from \{1e-4, 1e-3, 1e-2\}.
                \item Fine-tuning stage: We fine-tuned for 250 steps with a cosine annealing scheduler. We searched early stopping epoch from $\{10\%,20\%, 50\%\}$, size of the \textit{reweighting dataset} from $\{20, 100,500\}$ and fine-tuning learning rate from \{1e-2, 1e-3, 1e-4\}.
            \end{itemize}
            
    \item \textbf{\ours}:
            \begin{itemize}
                \item ERM stage: We used the same hyperparameter configuration with the aforementioned ERM model.
                \item Projector Training stage: We conducted a grid search on $\lambda$ in [1, 100] in units of 1.
                \item LLR stage: We used SGD as the optimizer with a batch size of 128 and trained the model for 50 epochs with a cosine annealing scheduler. 
                We searched learning rate from \{1e-4, 3e-4, 5e-4, 1e-3, 3e-3, 5e-3, 1e-2\} and set weight decay to 1e-4 without searching.
            \end{itemize}
\end{itemize}

\subsubsection{CelebA}
\begin{itemize}
    \item \textbf{ERM}:
    We used SGD as the optimizer with a batch size of 128 and trained the model for 50 epochs without any scheduler.
    We searched learning rate from \{1e-4, 1e-3, 3e-3, 1e-2\} and weight decay from \{1e-4, 1e-3, 1e-2\}.
    It is used for DFR and TLDR while SELF and AFR have their own ERM stage.

    {
    \item \textbf{Group-DRO}:
    We used SGD as the optimizer with a batch size of 128 and trained the model for 50 epochs without any scheduler.
    We searched learning rate and weight decay from a set of pairs \{(1e-5, 0.1), (1e-4, 1e-2), (1e-4, 1e-4)\} and $\eta_q$ from \{1e-4, 1e-3, 1e-2, 1e-1\}.
    }
    \item \textbf{DFR}:
            \begin{itemize}
                \item ERM stage: We used the same hyperparameter configuration with the aforementioned ERM model.
                \item LLR stage: We searched $\ell_1$ penalty from \{1e-2, 3e-2, 7e-2, 1e-1, 3e-1, 7e-1, 1.0\}.
            \end{itemize}
            
    \item \textbf{AFR}:
            \begin{itemize}
                \item ERM stage: We used SGD as the optimizer with a batch size of 128 and trained the model for 20 epochs with a cosine annealing scheduler.
                We searched learning rate from \{1e-4, 1e-3, 3e-3, 1e-2\} and weight decay from \{1e-4, 1e-3, 1e-2\}
                \item LLR stage: We trained the model for 1000 epochs.
                We searched $\gamma$ from 10 points linearly spaced between $[1, 3]$, learning rate from \{1e-2, 2e-2, 3e-2\} and $\lambda$ from \{1e-3, 1e-2, 1e-1\}.
            \end{itemize}
    \item \textbf{SELF}:
            \begin{itemize}
                \item Class-balanced ERM stage: We used SGD as the optimizer with a batch size of 100 and trained the model for 20 epochs with a cosine annealing scheduler.
                We searched the learning rate from \{1e-4, 1e-3, 3e-3, 1e-2\} and weight decay from \{1e-4, 1e-3, 1e-2\}.
                \item Fine-tuning stage: We fine-tuned for 250 steps with a cosine annealing scheduler. We searched early stopping epoch from 11 points linearly spaced between $[5\%, 50\%]$, size of the \textit{reweighting dataset} from $\{20, 100, 500\}$ and fine-tuning learning rate from \{1e-4, 1e-3, 1e-2\}.
            \end{itemize}
            
    \item \textbf{\ours}:
            \begin{itemize}
                \item ERM stage: We used the same hyperparameter configuration with the aforementioned ERM model.
                \item Projector Training stage: We conducted a grid search on $\lambda$ in [1, 10] in units of 1.
                \item LLR stage: We used SGD as the optimizer with a batch size of 128 and trained the model for 50 epochs with a cosine annealing scheduler.
                We searched learning rate from \{1e-4, 3e-4, 5e-4, 1e-3, 3e-3, 5e-3, 1e-2\} and set weight decay to 1e-4 without searching.
            \end{itemize}
\end{itemize}

\subsubsection{SpuCoAnimals}
\begin{itemize}

    \item \textbf{ERM}:
    We used SGD as the optimizer with a batch size of 128 and trained the model for 100 epochs without any scheduler.
    We searched learning rate from \{1e-4, 1e-3, 3e-3, 1e-2\} and weight decay from \{1e-4, 1e-3, 1e-2\}.
    It is used for DFR and TLDR while SELF and AFR have their own ERM stage.

    {
    \item \textbf{Group-DRO}:
    We used SGD as the optimizer with a batch size of 128 and trained the model for 100 epochs without any scheduler.
    We searched learning rate and weight decay from a set of pairs \{(1e-5, 1.0), (1e-4, 1e-1), (1e-3, 1e-4)\} and $\eta_q$ from \{1e-4, 1e-3, 1e-2, 1e-1\}.
    }

    \item \textbf{DFR}:
            \begin{itemize}
                \item ERM stage: We used the same hyperparameter configuration with the aforementioned ERM model.
                \item LLR stage: We searched $\ell_1$ penalty from \{1e-2, 3e-2, 7e-2, 1e-1, 3e-1, 7e-1, 1.0\}.
            \end{itemize}
            
    \item \textbf{AFR}:
            \begin{itemize}
                \item ERM stage: We used SGD as the optimizer with a batch size of 64 and trained the model for 50 epochs with a cosine annealing scheduler.
                We searched learning rate from \{1e-4, 1e-3, 3e-3, 1e-2\} and weight decay from \{1e-4, 1e-3, 1e-2\}.
                \item LLR stage: We trained the model for 500 epochs.
                We searched $\gamma$ from 10 points linearly spaced between $[1, 10]$, learning rate from \{1e-2, 2e-2, 3e-2\} and $\lambda$ from \{0, 1e-3, 1e-2, 1e-1\}.
            \end{itemize}
    \item \textbf{SELF}:
            \begin{itemize}
                \item Class-balanced ERM stage: We used SGD as the optimizer with a batch size of 64 and trained the model for 50 epochs with a cosine annealing scheduler.
                We searched the learning rate from \{1e-4, 1e-3, 3e-3, 1e-2\} and weight decay from \{1e-4, 1e-3, 1e-2\}.
                \item Fine-tuning stage: We fine-tuned for 250 steps with a cosine annealing scheduler. We searched early stopping epoch from $\{10\%, 20\%, 50\%\}$, size of the \textit{reweighting dataset} from $\{20, 100, 500\}$ and fine-tuning learning rate from \{1e-4, 1e-3, 1e-2\}.
            \end{itemize}
            
    \item \textbf{\ours}:
            \begin{itemize}
                \item ERM stage: We used the same hyperparameter configuration with the aforementioned ERM model.
                \item Projector Training stage: We conducted a grid search on $\lambda$ in [10000, 15000] in units of 100.
                \item LLR stage: We used AdamW as the optimizer with a batch size of 256 and trained the model for 200 epochs without any scheduler. 
                We searched learning rate from \{1e-1, 2e-1, 3e-1, 4e-1, 5e-1\} and set weight decay to 1e-4 without searching.
            \end{itemize}
\end{itemize}

\subsection{Details of \cref{tab:main_results} ($^\star$AFR) and \cref{fig:figure3_1}}
Except for the \cref{tab:main_results} ($^\star$AFR) and AFR on Waterbirds in \cref{fig:figure3_1} of the manuscript, we used the same hyperparameter search space for all ablation studies as stated in \cref{sec:hyper_search}.
The difference in \cref{tab:main_results} ($^\star$AFR) of the manuscript is due to lower learning rates are found to be not effective experimentally and the difference in \cref{fig:figure3_1} of the manuscript is due to a change of the configuration of the dataset.
The details are as follows.

\begin{itemize}
    \item \textbf{\cref{tab:main_results} ($^\star$AFR) on Waterbirds}:
        We only changed the learning rate search space for LLR stage to \{1e-1, 2e-1, 3e-1\}.
    \item \textbf{AFR on Waterbirds in \cref{fig:figure3_1}}:
        \begin{itemize}
            \item ERM stage: We used SGD as the optimizer with a batch size of 32 and trained the model for 50 epochs with a cosine annealing scheduler.
            We searched learning rate from \{1e-4, 1e-3, 3e-3, 1e-2\} and weight decay from \{1e-4, 1e-3, 1e-2\}.
            \item LLR stage: We trained the last linear layer for 1000 epochs. We searched $\gamma$ from 10 points linearly spaced between [1, 3], learning rate from \{1e-2, 3e-2, 5e-2\} and $\lambda$ from \{0, 1e-3, 1e-2, 1e-1\}.
        \end{itemize}
\end{itemize}

\subsection{Details of \cref{tab:rebut_table3}} 
\label{sec:exp_details_drml}
The hyperparameters used in \cref{tab:rebut_table3} of the manuscript are as follows: 
\begin{itemize}
    \item \textbf{ERM stage}: We used Adam as the optimizer with a batch size of 32 and only trained the last linear layer for 25 epochs without any scheduler. We used a learning rate of 1e-3 and a weight decay of 5e-4.
    \item \textbf{Fine-tuning stage}: We used Adam optimizer with a batch size of 32 and fine-tuned the last linear layer for 10 epochs without any scheduler. We searched learning rate from \{1e-3, 3e-3, 5e-3, 1e-2\} and did not used weight decay.
\end{itemize}
The best model was selected by validation loss on the image validation set, and the \textit{naive embeddings} in \cref{sec:ablation_umap} were used for all cases to align with the DrML's experimental setting.
The only difference is that DrML uses all 80 CLIP prompt templates to construct the text datasets for the fine-tuning.
In contrast, during the fine-tuning stage with TLDR's words, templates were randomly selected whenever each pair of words is fetched to reduce the computational cost.

\subsection{Details of \cref{sec:main_results_vit}}
\label{sec:exp_details_vit}

We reduced the batch size used for training ERM model used in ERM, DFR, \ours \ and experiments on post-hoc utilization of AFR and SELF due to memory constraints.
Also, there are slight modifications of search spaces of learning rate and $\lambda$ for \ours.
In addition, the search space of the learning rate for the experiment on post-hoc utilization of AFR is different from the experiment with ResNet-50.
The other details are the same with \cref{sec:hyper_search}.
\begin{itemize}
    \item \textbf{Reduced batch size}:
        We reduced the batch size for both CelebA and SpuCoAnimals to 64.
    \item \textbf{TLDR changes}:
        We changed the search space of learning rate to \{7e-5, 9e-5, 1e-4, 3e-4, 5e-4, 1e-3, 3e-3\} and batch size to 64 on Waterbirds and CelebA.
        In addition, we changed the search space of $\lambda$ on SpuCoAnimals to [200, 300] in units of 1.
    \item \textbf{AFR on Waterbirds in \cref{sec:main_results_vit}}:
        We used the learning rate search space for LLR stage as \{1e-2, 2e-2, 3e-2\}.
\end{itemize}

%% file: Table/data_configuration.tex
\parbox{.45\textwidth}{
\resizebox{0.45\textwidth}{!}{
\centering
\begin{tabular}{lcccccccc}
 \toprule
 \multicolumn{9}{c}{\textbf{Waterbirds}}\\ 
 \midrule
 \multicolumn{1}{c}{} & \multicolumn{4}{c}{Landbirds} & \multicolumn{4}{c}{Waterbirds}  \\
\cmidrule(r){2-5} \cmidrule(lr){6-9}
 Data Split & \multicolumn{2}{c}{Land}       &  \multicolumn{2}{c}{Water}         & \multicolumn{2}{c}{Land}       &  \multicolumn{2}{c}{Water}      \\
\midrule
Train      & \multicolumn{2}{c}{3498}   & \multicolumn{2}{c}{184 (4\%)}      & \multicolumn{2}{c}{56 (1\%)}   & \multicolumn{2}{c}{1057}       \\
Validation  & \multicolumn{2}{c}{467}   & \multicolumn{2}{c}{466}      & \multicolumn{2}{c}{133}   & \multicolumn{2}{c}{133}      \\
Test  & \multicolumn{2}{c}{2255}   & \multicolumn{2}{c}{2255}      & \multicolumn{2}{c}{642}   & \multicolumn{2}{c}{642}  
\\
\bottomrule
\end{tabular}
}
}
\hfill
\parbox{.45\textwidth}{
\resizebox{0.45\textwidth}{!}{
\centering
\begin{tabular}{lcccccccc}
 \toprule
 \multicolumn{9}{c}{\textbf{CelebA}}\\ 
 \midrule
 \multicolumn{1}{c}{} & \multicolumn{4}{c}{Non-blond} & \multicolumn{4}{c}{Blond}  \\
\cmidrule(r){2-5} \cmidrule(lr){6-9}
 Data Split & \multicolumn{2}{c}{Woman}       &  \multicolumn{2}{c}{Man}         & \multicolumn{2}{c}{Woman}       &  \multicolumn{2}{c}{Man}      \\
\midrule
Train      & \multicolumn{2}{c}{71629}   & \multicolumn{2}{c}{66874}      & \multicolumn{2}{c}{22880}   & \multicolumn{2}{c}{1387 (1\%)}       \\
Validation  & \multicolumn{2}{c}{8535}   & \multicolumn{2}{c}{8276}      & \multicolumn{2}{c}{2874}   & \multicolumn{2}{c}{182}          \\
Test  & \multicolumn{2}{c}{9767}   & \multicolumn{2}{c}{7535}      & \multicolumn{2}{c}{2480}   & \multicolumn{2}{c}{180}  \\
\bottomrule
\end{tabular}
}
}
\resizebox{\textwidth}{!}{
\centering
\begin{tabular}{lcccccccc}
\toprule
 \multicolumn{9}{c}{\textbf{SpuCoAnimals}}\\ 
 \midrule
 \multicolumn{1}{c}{} & \multicolumn{2}{c}{Landbirds} & \multicolumn{2}{c}{Waterbirds} & \multicolumn{2}{c}{Small Dogs} & \multicolumn{2}{c}{Big Dogs} \\
\cmidrule(r){2-3} \cmidrule(lr){4-5} \cmidrule(lr){6-7} \cmidrule(l){8-9}
 Data Split & Land       & Water      & Land        & Water       & Indoor      & Outdoor     & Indoor    & Outdoor\\
\midrule
Train      & 10000      & 500 (1.2\%)       & 500 (1.2\%)        & 10000       & 10000       & 500 (1.2\%)        & 500 (1.2\%)      & 10000 \\
Validation & 500        & 25         & 25          & 500         & 500         & 25          & 25        & 500 \\
Test & 500        & 500         & 500          & 500         & 500         & 500          & 500        & 500 \\
\bottomrule
\end{tabular}
}

%% file: Table/Nominority_waterbirds_configuration.tex
\begin{tabular}{lcccccccc}
 \toprule
 \multicolumn{9}{c}{\textbf{Waterbirds in \cref{fig:figure3_1}}}\\ 
 \midrule
 \multicolumn{1}{c}{} & \multicolumn{4}{c}{Landbirds} & \multicolumn{4}{c}{Waterbirds}  \\
\cmidrule(r){2-5} \cmidrule(lr){6-9}
 Data Split & \multicolumn{2}{c}{Land}       &  \multicolumn{2}{c}{Water}         & \multicolumn{2}{c}{Land}       &  \multicolumn{2}{c}{Water}      \\
\midrule
Train      & \multicolumn{2}{c}{3172}   & \multicolumn{2}{c}{522 (11\%)}      & \multicolumn{2}{c}{152 (3\%)}   & \multicolumn{2}{c}{949}       \\
Validation  & \multicolumn{2}{c}{793}   & \multicolumn{2}{c}{128}      & \multicolumn{2}{c}{37}   & \multicolumn{2}{c}{241}      \\
Test  & \multicolumn{2}{c}{2255}   & \multicolumn{2}{c}{2255}      & \multicolumn{2}{c}{642}   & \multicolumn{2}{c}{642}  
\\
\bottomrule
\end{tabular}

%% file: contents/Supp_templates.tex
\noindent \textbf{List of Prompt Templates for LLR}
\begin{lstlisting}[ language=Python]

    openai_imagenet_template = [
        lambda c: f"a bad photo of a {c}.",
        lambda c: f"a photo of many {c}.",
        lambda c: f"a sculpture of a {c}.",
        lambda c: f"a photo of the hard to see {c}.",
        lambda c: f"a low resolution photo of the {c}.",
        lambda c: f"a rendering of a {c}.",
        lambda c: f"graffiti of a {c}.",
        lambda c: f"a bad photo of the {c}.",
        lambda c: f"a cropped photo of the {c}.",
        lambda c: f"a tattoo of a {c}.",
        lambda c: f"the embroidered {c}.",
        lambda c: f"a photo of a hard to see {c}.",
        lambda c: f"a bright photo of a {c}.",
        lambda c: f"a photo of a clean {c}.",
        lambda c: f"a photo of a dirty {c}.",
        lambda c: f"a dark photo of the {c}.",
        lambda c: f"a drawing of a {c}.",
        lambda c: f"a photo of my {c}.",
        lambda c: f"the plastic {c}.",
        lambda c: f"a photo of the cool {c}.",
        lambda c: f"a close-up photo of a {c}.",
        lambda c: f"a black and white photo of the {c}.",
        lambda c: f"a painting of the {c}.",
        lambda c: f"a painting of a {c}.",
        lambda c: f"a pixelated photo of the {c}.",
        lambda c: f"a sculpture of the {c}.",
        lambda c: f"a bright photo of the {c}.",
        lambda c: f"a cropped photo of a {c}.",
        lambda c: f"a plastic {c}.",
        lambda c: f"a photo of the dirty {c}.",
        lambda c: f"a jpeg corrupted photo of a {c}.",
        lambda c: f"a blurry photo of the {c}.",
        lambda c: f"a photo of the {c}.",
        lambda c: f"a good photo of the {c}.",
        lambda c: f"a rendering of the {c}.",
        lambda c: f"a {c} in a video game.",
        lambda c: f"a photo of one {c}.",
        lambda c: f"a doodle of a {c}.",
        lambda c: f"a close-up photo of the {c}.",
        lambda c: f"a photo of a {c}.",
        lambda c: f"the origami {c}.",
        lambda c: f"the {c} in a video game.",
        lambda c: f"a sketch of a {c}.",
        lambda c: f"a doodle of the {c}.",
        lambda c: f"a origami {c}.",
        lambda c: f"a low resolution photo of a {c}.",
        lambda c: f"the toy {c}.",
        lambda c: f"a rendition of the {c}.",
        lambda c: f"a photo of the clean {c}.",
        lambda c: f"a photo of a large {c}.",
        lambda c: f"a rendition of a {c}.",
        lambda c: f"a photo of a nice {c}.",
        lambda c: f"a photo of a weird {c}.",
        lambda c: f"a blurry photo of a {c}.",
        lambda c: f"a cartoon {c}.",
        lambda c: f"art of a {c}.",
        lambda c: f"a sketch of the {c}.",
        lambda c: f"a embroidered {c}.",
        lambda c: f"a pixelated photo of a {c}.",
        lambda c: f"itap of the {c}.",
        lambda c: f"a jpeg corrupted photo of the {c}.",
        lambda c: f"a good photo of a {c}.",
        lambda c: f"a plushie {c}.",
        lambda c: f"a photo of the nice {c}.",
        lambda c: f"a photo of the small {c}.",
        lambda c: f"a photo of the weird {c}.",
        lambda c: f"the cartoon {c}.",
        lambda c: f"art of the {c}.",
        lambda c: f"a drawing of the {c}.",
        lambda c: f"a photo of the large {c}.",
        lambda c: f"a black and white photo of a {c}.",
        lambda c: f"the plushie {c}.",
        lambda c: f"a dark photo of a {c}.",
        lambda c: f"itap of a {c}.",
        lambda c: f"graffiti of the {c}.",
        lambda c: f"a toy {c}.",
        lambda c: f"itap of my {c}.",
        lambda c: f"a photo of a cool {c}.",
        lambda c: f"a photo of a small {c}.",
        lambda c: f"a tattoo of the {c}.",
    ]

\end{lstlisting}

%% file: contents/Supp_prompt.tex

\begin{lstlisting}[language=Python]
waterbirds_water_prompt_list = ["A photo of a waterbird in the ocean",
                                "A photo of a waterbird in the lake"]           
waterbirds_land_prompt_list  = ["A photo of a waterbird in the forest", 
                                "A photo of a waterbird in the broadleaf"]
landbirds_water_prompt_list  = ["A photo of a landbird in the ocean", 
                                "A photo of a landbird in the lake"]
landbirds_land_prompt_list   = ["A photo of a landbird in the forest", 
                                "A photo of a landbird in the broadleaf"]
                                
blond_male_prompt_list       = ["A photo of a man with blond hair"]
non_blond_male_prompt_list   = ["A photo of a man with dark hair"]
blond_female_prompt_list     = ["A photo of a woman with blond hair"]
non_blond_female_prompt_list = ["A photo of a woman with dark hair"]

bigdog_outdoor_prompt_list   = ["A photo of a big dog in the outdoor"]
bigdog_indoor_prompt_list    = ["A photo of a big dog in the indoor"]
smalldog_outdoor_prompt_list = ["A photo of a small dog in the outdoor"]
smalldog_indoor_prompt_list  = ["A photo of a small dog in the indoor"]

\end{lstlisting}